\documentclass{article}
\PassOptionsToPackage{numbers}{natbib}
\usepackage[final]{neurips_distshift_2022}

\usepackage[utf8]{inputenc} % allow utf-8 input
\usepackage[T1]{fontenc}    % use 8-bit T1 fonts
\usepackage{hyperref}       % hyperlinks
\usepackage{url}            % simple URL typesetting
\usepackage{booktabs}       % professional-quality tables
\usepackage{amsfonts}       % blackboard math symbols
\usepackage{nicefrac}       % compact symbols for 1/2, etc.
\usepackage{microtype}      % microtypography
\usepackage[dvipsnames]{xcolor}         % colors

\usepackage{graphicx}
\usepackage[normalem]{ulem}
\usepackage{amsmath}
\usepackage{bm}
\usepackage{bbm}
\usepackage{multirow}
\usepackage{booktabs}
\usepackage{amssymb}
\usepackage[nolist,nohyperlinks]{acronym}
\usepackage{arydshln}
\usepackage{enumitem}
\usepackage{xspace}
\usepackage{float}

\usepackage{pgfplots}
\usepackage{wrapfig}
\pgfplotsset{compat=1.16}
\pgfplotsset{every tick label/.append style={font=\tiny}}
\usepackage{pgfplotstable}
\usetikzlibrary{pgfplots.groupplots}
\usepackage{hyperref}

\newcommand*{\EOP}{\hfill\ensuremath{\square}}%

\newcommand{\MMD}{\mbox{MMD}}

\newtheorem{definition}{Definition}

\newtheorem{lemma}{Lemma}

\def\mb#1{\mathbf{#1}}

\newcommand{\leakage}{\emph{transfer leakage}\xspace}
\newcommand{\pleakage}{\emph{pseudo transfer leakage}\xspace}
\newcommand{\Leakage}{\emph{Transfer leakage}\xspace}
\newcommand{\Pleakage}{\emph{Pseudo transfer leakage}\xspace}

\title{A Closer Look at Novel Class Discovery \\ from the Labeled Set}

\author{%
Ziyun Li \\
Hasso Plattner Institute\\
\texttt{ziyun.li@hpi.de} \\
\And
Jona Otholt \\
Hasso Plattner Institute\\
\texttt{jona.Otholt@hpi.de} \\
\And
Ben Dai\thanks{Corresponding author}\\
Chinese University of Hong Kong \\
\texttt{bendai@cuhk.edu.hk } \\
\AND
Di Hu \\
Renmin University of China\\
\texttt{dihu@ruc.edu.cn} \\
\And
Christoph Meinel \\
Hasso Plattner Institute\\
\texttt{christoph.meinel@hpi.de} \\
\And
Haojin Yang$^*$ \\
  Hasso Plattner Institute\\
  \texttt{haojin.yang@hpi.de} \\
}

\begin{document}
\maketitle
\begin{abstract}
    Novel class discovery (NCD) is to infer novel categories in an unlabeled set using prior knowledge of a labeled set comprising diverse but related classes.
    Existing research focuses on using the labeled set methodologically and little on analyzing it.
    In this study, we take a closer look at NCD from the labeled set and focus on two questions:
    (i) Given an unlabeled set, \textit{what labeled set best supports novel class discovery?}
    (ii) A fundamental premise of NCD is that the labeled set must be related to the unlabeled set, but \textit{how can we measure this relation?}
    For (i), we propose and substantiate the hypothesis that NCD could benefit from a labeled set with high semantic similarity to the unlabeled set.
    Using ImageNet's hierarchical class structure, we create a large-scale benchmark with variable semantic similarity across labeled/unlabeled datasets.
    In contrast, existing NCD benchmarks ignore the semantic relation.
    For (ii), we introduce a mathematical definition for quantifying the semantic similarity between labeled and unlabeled sets.
    We utilize this metric to validate our established benchmark and demonstrate it highly corresponds with NCD performance.
    Furthermore, without quantitative analysis, previous works commonly believe that label information is always beneficial.
    However, our experimental results counterintuitively show that using labels may lead to suboptimal outcomes in low-similarity settings.

\end{abstract}

\section{Introduction}
\label{sec:intro}

% Deep models are capable of properly identifying and clustering classes that are present in the training set (i.e., known/seen classes), matching or surpassing human performance. 
% However, they lack reliable extrapolation capacity when confronted with novel classes while humans can easily recognize the unseen categories.
% This motivated researchers to develop a challenge termed novel class discovery (NCD)~\cite{han2019learning, chi2021meta, han2021autonovel, zhong2021neighborhood}, 
% with the goal of discovering novel classes in an unlabeled dataset by leveraging knowledge from a labeled set, which contains related but disjoint classes.

Deep models are capable of identifying and clustering classes that are present in the training set (i.e., known/seen classes), matching or surpassing human performance. 
However, they lack reliable extrapolation capacity when confronted with novel classes, while humans can easily recognize the unseen categories.
This encouraged researchers to establish a challenge termed novel class discovery (NCD)~\cite{han2019learning, chi2021meta, han2021autonovel, zhong2021neighborhood}, 
to identify new classes in an unlabeled dataset by utilizing information from a labeled set containing similar but disjoint classes.

Currently, most NCD research takes place at the method level, focusing on better utilizing the labeled set.
Though the labeled set is essential, there is a less in-depth analysis of the labeled set itself.
% The sole study on this area is \cite{chi2021meta}, which finds that NCD is theoretically solvable 
% when labeled and unlabeled data share similar high-level semantic features.
This lack of understanding about a crucial aspect of NCD illustrates the necessity to explore it from the labeled set's perspective.
Thus, our paper concentrates on two core questions:
First, given a specific unlabeled set, \emph{what kind of labeled set can best support novel class discovery?}
Second, an essential premise of NCD is that the labeled set should be related to the unlabeled set, but \emph{how can we measure this relation?}
Based on the preceding questions, we also give insights into the importance of labeled information in NCD.

Regarding the first question, we intuitively expect that labeled sets with higher semantic similarity can provide more beneficial knowledge
while the number of categories and pictures is fixed.
In contrast, existing works solely use the number of labeled/unlabeled classes and images to determine NCD difficulty, 
e.g., \cite{fini2021unified, chi2021meta}, and disregard semantic similarity.
We first verify our assumption on multiple pairs 
of labeled and unlabeled sets with varying semantic similarity 
and under multiple baselines \cite{han2019learning, han2021autonovel, fini2021unified, macqueen1967some}.
Then, we establish a new benchmark with multiple semantic similarity levels using ImageNet's hierarchical semantic information,
and more details can be found in Section~\ref{sec:higher greater}.

Second, an essential premise of NCD is that leveraging the information of the disjoint but related labeled set improves performance on the unlabeled data.
A prior work \cite{chi2021meta} points out that NCD is theoretically solvable when labeled set and unlabeled set 
share high-level semantic features yet without proposing any quantitative analysis.
This inspires the following questions:
How closely related do the sets need to be for NCD to work?
How can we measure the semantic relatedness between labeled and unlabeled sets?
%Such a metric would also be beneficial to judge the difficulty of our new benchmark.
Motivated by these questions, we propose a semantic similarity metric, called \leakage.
Specifically, \leakage quantifies how much information we can leverage from the labeled dataset to help improve the performance of the unlabeled dataset, 
and more details are provided in Section~\ref{sec:transfer_leakage}.

Furthermore, we observe that labeled information may lead to sub-optimal results, contrary to the commonly held belief that labeled information is always beneficial for NCD tasks.
% We modify the current state-of-the-art NCD approach~\cite{fini2021unified} to work without labels and show that this version can outperform the standard method in cases where the semantic similarity between the labeled and unlabeled sets is low.
% We conduct experiments on the labeled set using the current state-of-the-art (SOTA) approach \cite{fini2021unified}, one with labeled information (standard) and one without labeled information.
% We observe that the using images-only information can outperform image-label pairs in cases where the semantic similarity between the labeled and unlabeled sets is low.
However, it is hard to decide whether to use labeled supervised knowledge or self-supervised knowledge without labels.
Thus, we provide two concrete solutions.
(i) \pleakage, a practical reference for what sort of data we intend to employ.
(ii) A straightforward method, which smoothly combines supervised and self-supervised knowledge from the labeled set and achieves 3\% and 5\% improvement in both CIFAR100 and ImageNet compared to SOTA. For further information, see Section \ref{sec:observation}.

% Thirdly, it is commonly believed that the supervised privileged knowledge is beneficial for NCD.
% 而我们的empirical 实验得出的结论有所不同。
% 我们分别在state-of-the-art uno 以及 typical rs上进行验证on cifar100, and our benchmark which build on imagenet.
% 我们做了三个对比实验：
% 1）only use unlabeled set
% 2) unlabeled set and labeled set's image 
% 3) unlabeled set and labeled set
% finding:
% 1.labeled set's image always helpful
% 2. supervised knowledge 在high semantic下有作用，然而在low semantic下，甚至有可能导致更worse的效果。
% 基于这个worse的效果，我们做进一步研究，supervised到底提供的是什么信息?
% 我们认为是rule规则，如果分类不一致，效果会变差，。。。。
% Thirdly, it is commonly believed that supervised knowledge is beneficial for discovering unknown categories.
% Thirdly, 
% %有标签的数据集比无标签难收集，不管是known 还是 unknown class
% %how about if we can‘t find 有标签且相关的数据集，can we borrow the knowledge from related known classes or high noise rate unlabeled set.
% %
% Does this hold true in any NCD settings, and if so, how significant is supervised knowledge to NCD's performance?
% According to the above hypothesis, we conduct experiments under the following data settings on CIFAR100 and our proposed benchmark:
% (i) only the unlabeled set without the labeled set
% (ii) the unlabeled set and the labeled set's images, but not the labeled set's labels;
% (iii) the unlabeled set and the labeled set's images with labels;
%%% 我们已经知道有标签的相关
%

We summarize our contributions as follows:
(i) We establish a comprehensive and large benchmark with varying degrees of difficulty on ImageNet and thoroughly justify the assumption that semantic similarity is a significant factor influencing NCD performance.
(ii) We introduce a mathematical definition for evaluating the semantic similarity between labeled and unlabeled sets and validate it under CIFAR100 and ImageNet.
(iii) We observe counterintuitive results - labeled information may lead to suboptimal performance and propose two practical applications, which achieve 3\% and 5\% improvement in both CIFAR100 and ImageNet compared to SOTA.

\begin{figure}[t]
    % \vspace{-0.0cm}
    \centering
    \includegraphics[width=1.0\linewidth]{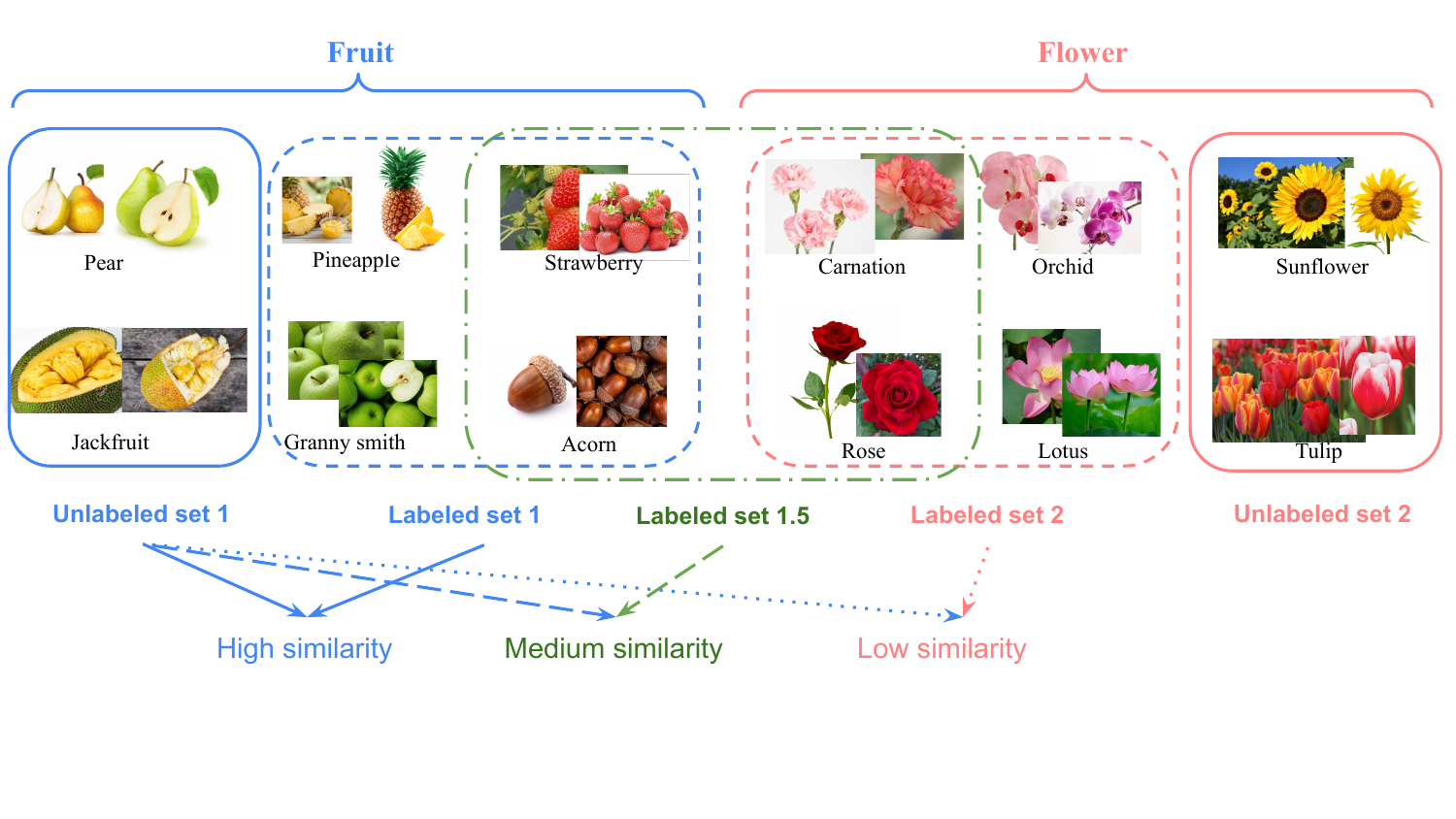}
    \vspace{-1.95cm}
    %\setlength{\tabcolsep}{2.8pt} % Default value: 6pt
    % \fontsize{9.1pt}{9.1pt}\selectfont
    \caption{Illustration of how we construct the benchmark with varying levels of semantic similarity.
    Unlabeled set $U_{1}$ and labeled set $L_{1}$ are from the same superclass (fruit), whereas unlabeled set $U_{2}$ and labeled set $L_{2}$ belong to another superclass (flower).
    Labeled set $L_{1.5}$ is composed of half of $L_{1}$ and half of $L_{2}$.
    If both the labeled and unlabeled classes are derived from the same superclass, i.e. ($U_{1}$, $L_{1}$) and ($U_{2}$, $L_{2}$), we consider them a \textbf{high semantic similarity} split.
    In contrast, ($U_{1}$, $L_{2}$) and ($U_{2}$, $L_{1}$) are \textbf{low semantic similarity} splits, since the labeled and unlabeled classes are derived from distinct superclasses.
    In addition, we consider ($U_{1}$, $L_{1.5}$) and ($U_{2}$, $L_{1.5}$) to have \textbf{medium semantic similarity} because half of $L_{1.5}$ share the same superclass as $U_{1}$.}
    \vspace{-3mm}
    \label{fig:splits}
    \end{figure}

\section{Assumption and Proposed Benchmarks}
%  {better title, not ready for review}}
\label{sec:higher greater}
In this section, we address the first question:
\emph{given a specific unlabeled set, what kind of labeled set can best support novel class discovery?}
We first assume that {higher semantic similarity labeled sets can provide greater help compared to less similar labeled sets} 
when the number of categories and images in the labeled sets are fixed.
However, existing benchmarks were created based on the number of categories(e.g., \cite{fini2021unified}) and images(e.g.,\cite{chi2021meta}) 
without considering semantic similarity between the two sets.
Thus, we propose a new benchmark based on the ENTITY-30 task\cite{santurkar2020breeds} 
including three different semantic similarity levels (high, medium and low) 
by leveraging the underlying hierarchy,
which contains 240 ImageNet classes in total, with 30 superclasses and 8 subclasses for each superclasses.
For our benchmark, we use these classes to create NCD tasks with 90 labeled and 30 unlabeled classes each.
Lastly, our hypothesis is verified on our benchmark (\autoref{tab:imagenet_benchmark}) and CIFAR100 (\autoref{tab:cifar100_benchmark}). 
The results demonstrate that the most similar labeled set achieves the highest performance, followed by the medium and the least similar set.
Further details are provided in the Appendix \ref*{sec:detail-of-benchmark}.

\begin{table}[]
    \centering
      \fontsize{9.3pt}{9.3pt}\selectfont
        \caption{
          Comparison of different semantic similarity settings in our proposed benchmark. 
          $L_{1}$ is closely related to $U_{1}$ and $L_{2}$ is highly related to $U_{2}$.
             The third labeled set $L_{1.5}$ is constructed from half of $L_{1}$ and half of $L_{2}$, so in terms of similarity it is in between $L_{1}$ and $L_{2}$.
             For all splits we report the mean and the standard deviation of the clustering accuracy across multiple NCD baselines.
             }
        \label{tab:imagenet_benchmark}
        \vspace{-2mm}
        \begin{tabular}{@{}lcccccc@{}}
        \toprule
        \multirow{2}{*}{Methods}               & \multicolumn{3}{c}{Unlabeled set $U_{1}$}     & \multicolumn{3}{c}{Unlabeled set $U_{2}$}     \\ 
                                        \cmidrule(l){2-4}                          \cmidrule(l){5-7} 
                                        % & close ($U_{1}$, $L_{1}$) & middle ($U_{1}$, $L_{1.5}$) & far ($U_{1}$, $L_{2}$)  & far ($U_{2}$, $L_{1}$)  & middle ($U_{2}$, $L_{1.5}$) & close ($U_{2}$, $L_{2}$) \\ \midrule
                                        % & ($U_{1}$, $L_{1}$) & ($U_{1}$, $L_{1.5}$) & ($U_{1}$, $L_{2}$)  & ($U_{2}$, $L_{1}$)  & ($U_{2}$, $L_{1.5}$) & ($U_{2}$, $L_{2}$) \\ \midrule
                                        & $L_{1}$ - high     & $L_{1.5}$ - medium     & $L_{2}$ - low       & $L_{1}$ - low       & $L_{1.5}$ - medium    & $L_{2}$ - high      \\ 
                                        \midrule
        K-means \cite{macqueen1967some} & \textbf{41.1 $\pm$ 0.4} & 30.2 $\pm$ 0.4 & 23.3 $\pm$ 0.2 & 21.2 $\pm$ 0.2 & 29.8 $\pm$ 0.4 & \textbf{45.0 $\pm$ 0.4}  \\
        DTC \cite{han2019learning}      & \textbf{43.3 $\pm$ 1.2} & 35.6 $\pm$ 1.3 & 32.2 $\pm$ 0.8 & 21.3 $\pm$ 1.2 & 15.3 $\pm$ 1.5 & \textbf{29.0 $\pm$ 0.8}  \\
        RS \cite{han2020automatically}  & \textbf{55.3 $\pm$ 0.4} & {50.3 $\pm$ 0.9} &	53.6 $\pm$ 0.6    & 48.1 $\pm$ 0.4 &	50.9 $\pm$ 0.6 &	\textbf{55.8 $\pm$ 0.7}      \\
         {NCL} \cite{zhong2021neighborhood}  &  {
        \textbf{75.1 $\pm$ 0.8}} &  {74.3 $\pm$ 0.4} &	 {71.6 $\pm$ 0.4}    &  {61.3 $\pm$ 0.1} &	 {70.5 $\pm$ 0.8} &	 {\textbf{75.1 $\pm$ 1.2}}      \\
        UNO \cite{fini2021unified}      & \textbf{83.9 $\pm$ 0.5} & 81.0 $\pm$ 0.5 & 77.2 $\pm$ 0.8 & 77.5 $\pm$ 0.7 & 82.0 $\pm$ 1.7 & \textbf{88.4 $\pm$ 1.2}  
        \\ \bottomrule
        \end{tabular}
        \vspace{-5mm}
    \end{table}

\section{Quantifying Semantic Similarity }
\label{sec:transfer_leakage}
% To the best of our knowledge, \textit{how to measure the semantic similarity between the labeled set and the unlabeled set remains unsolved.}
% Thus, we focus on this core question.
% In this section,
% we first present a formal framework of NCD, then introduce a mathematical definition on measuring the semantic similarity between labeled and unlabeled sets.
\subsection{NCD Framework}
We denote $(\mb{X}_l, Y_l)$ and $(\mb{X}_u, Y_u)$ as random samples under the \emph{labeled/unlabeled probability measures} $\mathbb{P}_{\mb{X}, Y}$ and $\mathbb{Q}_{\mb{X}, Y}$, respectively.
$\mb{X}_l \in \mathcal{X}_l \subset \mathbb{R}^d$ and $\mb{X}_u \in \mathcal{X}_u \subset \mathbb{R}^d$ are the labeled/unlabeled feature vectors, $Y_l \in \mathcal{C}_l$ and $Y_u \in \mathcal{C}_u$ are the true labels of labeled/unlabeled data, where $\mathcal{C}_l$ and $\mathcal{C}_u$ are the label sets under the labeled and unlabeled probability measures $\mathbb{P}_{\mb{X}, Y}$ and $\mathbb{Q}_{\mb{X}, Y}$, respectively.
Given a labeled set $\mathcal{L}_n = (\mb{X}_{l,i}, Y_{l,i})_{i=1}^n$ independently drawn from 
the labeled probability measure $\mathbb{P}_{\mb{X},Y}$, and an unlabeled dataset 
$ \mathcal{U}_m = (\mb{X}_{u,i})^m_{i=1}$ independently drawn from the unlabeled probability measure 
$\mathbb{Q}_{\mb{X}_u}$, our primary goal is to predict $Y_{u,i}$ given $\mb{X}_{u,i}$,  
{where $Y_{u,i}$ is the label of the $i$-th unlabeled sample $\mb{X}_{u,i}$}. 
% We now give a general definition of NCD.

\begin{definition}[Novel class discovery]
	Let $\mathbb{P}_{\mb{X}_l,Y_l}$ be a labeled probability measure on $\mathcal{X}_l \times \mathcal{C}_l$, and $\mathbb{Q}_{\mb{X}_u,Y_u}$ be an unlabeled probability measure on $\mathcal{X}_u \times \mathcal{C}_u$, with $\mathcal{C}_u \cap \mathcal{C}_l = \emptyset$. Given a labeled dataset $\mathcal{L}_n$ sampled from $\mathbb{P}_{\mb{X}_l, Y_l}$ and an unlabeled dataset $\mathcal{U}_m$ sampled from $\mathbb{Q}_{\mb{X}_u}$, novel class discovery aims to predict the labels of the unlabeled dataset based on $\mathcal{L}_n$ and $\mathcal{U}_m$.
\end{definition}

\subsection{Transfer Leakage}
\label{lab:transfer leakage}
% Here, we address a central question: \textit{How to measure the similarity between the labeled set under $\mathbb{P}$ and the unlabeled set under $\mathbb{Q}$?}

% Note that it is critical to measure the similarity between $\mathbb{P}^l$ and $\mathbb{P}^u$.
% As mentioned in Section~\ref{sec:higher greater}, different pairs of labeled/unlabeled datasets lead to significant different predictive performance.
% A highly related (semantically similar) labeled set can provide more beneficial information for discovering novel classes.
% This raises a central question: \textit{How to measure the similarity between the labeled set under $\mathbb{P}$ and the unlabeled set under $\mathbb{Q}$?} 
%
% As noted in Section, different pairs of labeled/unlabeled datasets lead to varying predictive performance.
% Highly related (semantically comparable) labeled sets can help find novel classes.
% It poses a key question. textitHow to measure similarity between labeled set $mathbbP$ and unlabeled set $mathbbQ$?
%
% This raises a central question:
% \begin{center}
% \vspace{-0cm}
% \textit{
% How to measure the similarity between the labeled set under $\mathbb{P}$ and the unlabeled set under $\mathbb{Q}$?
% % How much can the labeled dataset $\mathcal{L}_n$ under $\mathbb{P}^l$ help to predict the label under $\mathbb{P}^u$?
% } 
% \vspace{-0cm}
% \end{center}
%To the best of our knowledge, the question of how to measure the semantic similarity between the labeled and unlabeled set in NCD remains unsolved.

% Before discussing the semantic similarity metric, 
We begin with introducing Maximum Mean Discrepancy (MMD) \cite{gretton2012kernel}, 
which is used to measure the discrepancy of two distributions. 
For example, the discrepancy of two random variables 
$\mb{Z} \sim \mathbb{P}_{\mb{Z}}$ and $\mb{Z}' \sim \mathbb{P}_{\mb{Z}'}$ is defined as:
$\MMD_\mathcal{H} \big( \mathbb{P}_{\mb{Z}}, \mathbb{P}_{\mb{Z}'} \big) := \sup_{ \|h\|_{\mathcal{H}} \leq 1 } \Big( \mathbb{E}\big( h( \mb{Z}) \big) - \mathbb{E}\big( h( \mb{Z}') \big) \Big)$
% \begin{align}
% 	\label{def:mmd}
% 	\MMD_\mathcal{H} \big( \mathbb{P}_{\mb{Z}}, \mathbb{P}_{\mb{Z}'} \big) := \sup_{ \|h\|_{\mathcal{H}} \leq 1 } \Big( \mathbb{E}\big( h( \mb{Z}) \big) - \mathbb{E}\big( h( \mb{Z}') \big) \Big),
% \end{align}
where $\mathcal{H}$ is a class of functions $h: \mathcal{X}_u \to \mathbb{R}$, which is specified as a reproducing kernel Hilbert Space (RKHS) associated with a continuous kernel function $K(\cdot, \cdot)$.
% From \eqref{def:mmd}, when $\MMD$ is large, the distributions between $\mb{Z}$ and $\mb{Z}'$ appear dissimilar.
%

% it provides a relatively large possibility to find a good transformation on $\mb{X}_u$ discriminating labels in an unsupervised manner. 
% For multi-class discrepancy, we mimic the one-vs-rest nature by taking the minimum discrepancy over all rest classes, which takes the discrepancy to the closest class. 
In NCD, the unlabeled dataset utilizes the conditional probability $\mathbb{P}_{Y_l | \mb{X}_l }$ (usually presented by a pretrained neural network) from a labeled dataset. 
For example, if the distributions of $\mathbb{P}_{Y_l | \mb{X}_l = \mb{X}_u}$ under $Y_u = c$ and $Y_u = c'$ are significantly different, then its overall distribution discrepancy is large, yielding that more information can be leveraged in NCD.
On this ground, we use MMD to quantify the discrepancy of the labeled probability measure $\mathbb{P}_{Y_l|\mb{X}_l}$ on $\mb{X}_u$ under the unlabeled probability measure $\mathbb{Q}$, namely \leakage.

\begin{definition}[Transfer leakage]
	\label{def:T-leak}
	The \leakage of NCD prediction under $\mathbb{Q}$ based on the labeled conditional probability $\mathbb{P}_{Y_l|\mb{X}_l}$ is
	\begin{equation}
	\text{T-Leak}( \mathbb{Q}, \mathbb{P}) = \mathbb{E}_{\mathbb{Q}} \Big( \MMD_{\mathcal{H}}^2 \big( \mathbb{Q}_{ \mb{p}(\mb{X}_u) | Y_u }, \mathbb{Q}_{\mb{p}(\mb{X}_u') \mid Y_u'} \big) \Big),
	\end{equation}
	where $(\mb{X}_u, Y_u), (\mb{X}_u', Y_u') \sim \mathbb{Q}$ are independent copies, the expectation $\mathbb{E}_{\mathbb{Q}}$ is taken with respect to $Y_u$ and $Y_u'$ under $\mathbb{Q}$, and $\mb{p}(\mb{x})$ is the conditional probability under $\mathbb{P}_{Y_l|\mb{X}_l}$ on an unlabeled data $\mb{X}_u = \mb{x}$, which is defined as
	$
	\mb{p}(\mb{x}) = \big(\mathbb{P}\big( Y_l=c \mid \mb{X}_l = \mb{x} \big) \big)^{\intercal}_{c \in \mathcal{C}_l}.
	$
\end{definition}
To summarize, \leakage measures the overall discrepancy of $\mb{p}(\mb{X}_u)$ under different new classes of the unlabeled measure $\mathbb{Q}$, which indicates the informative leakage from $\mathbb{P}$ to $\mathbb{Q}$. 
Next, we give a finite sample estimate of \leakage. 
{Given an estimated probability $\widehat{\mathbb{P}}_{Y_l|\mb{X}_l}$ and an evaluation dataset $(\mb{x}_{u,i}, y_{u,i})_{i=1}^m$ under $\mathbb{Q}$}, we assess $\mb{x}_{u,i}$ on $\widehat{\mathbb{P}}_{Y_l | \mb{X}_l}$ as $\widehat{\mb p}(\mb{x}_{u,i}) = \big(\widehat{\mathbb{P}} \big( Y_l=c | \mb{X}_l = \mb{x}_{u,i} \big) \big)^{\intercal}_{c \in \mathcal{C}_u}$, then empirical \leakage is computed as:
\begin{align}
	\label{eqn:emp_transfer_leak}
	\widehat{\text{T-Leak}}( \mathbb{Q}, \mathbb{P} ) &= \sum_{c, c' \in \mathcal{C}_u; c \neq c'} \frac{|\mathcal{I}_{u,c}| |\mathcal{I}_{u,c'}|}{m(m-1)} \widehat{\MMD}^2_{\mathcal{H}} \big( \mathbb{Q}_{ \widehat{\mb{p}}(\mb{X}_u) | Y_u = c }, \mathbb{Q}_{\widehat{\mb{p}}(\mb{X}_u') \mid Y_u' = c} \big),
\end{align}
where the equality follows from the fact that $\MMD_{\mathcal{H}}^2 \big( \mathbb{Q}_{ \mb{p}(\mb{X}_u) | Y_u = c }, \mathbb{Q}_{\mb{p}(\mb{X}_u') \mid Y_u' = c} \big) = 0$.
$\mathcal{I}_{u,c} = \{ 1 \leq i \leq m: y_{u,i} = c \}$ is the index set of unlabeled data with $y_{u,i} = c$, and $\widehat{\MMD}^2_{\mathcal{H}}$ is defined in the Appendix.
%\red{where the equality follows from the fact that $\MMD_{\mathcal{H}}^2 \big( \mathbb{Q}_{ \mb{p}(\mb{X}_u) | Y_u = c }, \mathbb{Q}_{\mb{p}(\mb{X}_u') \mid Y_u' = c} \big) = 0$.}
We use the proposed $\widehat{\text{T-Leak}}$ to quantify the difficulty of NCD in various combination of labeled and unlabeled sets. It is worth noting that the proposed \textit{transfer leakage} and its empirical evaluation depend only on the labeled/unlabeled datasets, and it remains the same, no matter what NCD method we use. 
In addition, we provide \pleakage (Appendix \ref{sec:p_tl}), a practical evaluation of the similarity between the classified and unlabeled sets.
\leakage utilizes the true label $Y_u$, but the \pleakage utilizes the pseudo label obtained from clustering methods (e.g., $k$-means) on the representations.
Further information is available in Appendix \ref{sec:detail-of-tleak}.

\section{Supervised Knowledge may be Harmful}
% \section{Initial Findings on the Role of Labels in NCD}
\label{sec:observation}
% \subsection{Background}

% The motivation behind NCD is that prior supervised knowledge from labeled data can help improve the clustering of the unlabeled set.
% However, we have counterintuitive results:
% supervised information from a labeled set may result in suboptimal outcomes as compared to using exclusively self-supervised knowledge.

The motivation behind NCD is that supervised knowledge from labeled data can enhance unlabeled data clustering.
Counterintuitively, we observe that supervised information from a labeled set 
may result in suboptimal outcomes compared to exclusively self-supervised knowledge.
Further information is provided in Appendix \ref{sec: appendix-practical}.

% And we conduct a fine-grained experiments on studying the gain from 
% the whole labeled set $(\mb{X}_l, Y_l)$, 
% Concretely, in NCD, given a labeled set $(\mb{X}_l, Y_l)$ and an unlabeled set $\mb{X}_u$,
\vspace*{-0.25cm}
\subsection{Experiments and Results}
To investigate, we conduct experiments in the following settings:
(1) Using the unlabeled set, $\mb{X}_u$;
(2) Using the unlabeled set and the labeled set's images without labels, $\mb{X}_u+ \mb{X}_l$; 
(Even without labels, self-supervised learning can extract the knowledge from labeled set's image.)
(3) Using the unlabeled set and the whole labeled set, $\mb{X}_u+ (\mb{X}_l, Y_l)$, (i.e., standard NCD).
As suggested in Table \ref{tab:counterintuitive-new}, NCD performance is consistently improved by incorporating more images (without labels) from a labeled set,  around 10\% on CIFAR100 and 6\%-18\% in our benchmark.
By comparing (2) and (3), we can isolate the impact of the labels.
Unexpectedly, on CIFAR100-50 and ImageNet with low semantic similarity, (3) performs around 2 - 8\% worse than (2), 
yielding that ``low-quality" supervised information may hurt NCD performance.
% The comparison between (1) and (2) as well as (2) and (3) allows us to further disentangle the performance gain according to the components of the labeled set.

\begin{table}[H]
    \centering
    \caption{Comparison of different data settings on CIFAR100 and our proposed benchmark.
    We present clustering mean and standard error on SOTA method, UNO.
    (1) uses only the unlabeled set, whereas (2) uses both the unlabeled set and the labeled set's images without labels.
    (3) represents the standard NCD setting, i.e., using the unlabeled set and the whole labeled set.
    Counterintuitively, in CIFAR100-50 and low similarity case of our benchmark, (2) can get greater performance than (3).
    }
    \label{tab:counterintuitive-new}
    \vspace{-2.mm}
    \begin{tabular}{@{}lccccc@{}}
    
          \toprule
    % \multicolumn{1}{l}{}       & \multicolumn{1}{l}{} &         & High semantic  & Middle semantic  & Low semantic  \\
    % \multirow{2}{*}{UNO}           &            & CIFAR100-50       & ($L_{1}$-$U_{1}$)      &    ($L_{1.5}$-$U_{1}$)        &    ($L_{2}$-$U_{1}$)     \\
     \multirow{2}{*}{Setting} & \multirow{2}{*}{CIFAR100-50}  & \multicolumn{2}{c}{Unlabeled set $U_{1}$}     & \multicolumn{2}{c}{Unlabeled set $U_{2}$}  \\
     \cmidrule(l){3-4} \cmidrule(l){5-6} 
     & &  $L_{1}$ - high &  $L_{2}$ - low & $L_{1}$ - low &  $L_{2}$ - high\\
    \midrule
    % \multirow{3}{*}{Avg head}  & (1) $\mb{X}_u$                 & 54.2 $\pm$ 0.3          & 69.2 $\pm$ 0.7          & 69.2 $\pm$ 0.7          & 69.2 $\pm$ 0.7             \\
    %                           & (2) $\mb{X}_u+ \mb{X}_l$        & \textbf{63.4 $\pm$ 0.4}          & 74.9 $\pm$ 0.3          & 77.6 $\pm$ 0.9          & \textbf{77.9 $\pm$ 1.1}   \\
    %                           & (3) $\mb{X}_u+ (\mb{X}_l, Y_l)$         & 61.7 $\pm$ 0.3          & \textbf{81.7 $\pm$ 1.0} & \textbf{80.3 $\pm$ 0.4} & 74.6 $\pm$ 0.3             \\
    %                           \midrule
                                 (1) $\mb{X}_u$                 & 54.9 $\pm$ 0.4          & 70.5 $\pm$ 1.2                   & 70.5 $\pm$ 1.2     &  71.9 $\pm$ 0.3   &  71.9 $\pm$ 0.3\\
                               (2) $\mb{X}_u+ \mb{X}_l$          & \textbf{64.1 $\pm$ 0.4} & 79.6 $\pm$ 1.1                    & \textbf{80.3 $\pm$ 0.3} &  \textbf{85.3 $\pm$ 0.5} & 89.2 $\pm$ 0.3 \\
                               (3) $\mb{X}_u+ (\mb{X}_l, Y_l)$        & 62.2 $\pm$ 0.2          & \textbf{83.9 $\pm$ 0.6}  & 77.2 $\pm$ 0.8     &  77.5 $\pm$ 0.7  &   88.3  $\pm$ 1.1 \\
      \bottomrule
    \end{tabular}
    \vspace{-3mm}
    \end{table}

\vspace*{-0.3cm}
\subsection{Practical Applications}
\label{sec: practical}
%As illustrated in Table \ref{tab:counterintuitive} and Table \ref{tab:counterintuitive u2}, 
As shown above, supervised knowledge from the labeled set may cause damage, 
but it's difficult to determine whether to utilize it or self-supervised knowledge.
Therefore, we offer two concrete solutions, a practical metric (i.e., \pleakage) and a straightforward method.
\paragraph{Supervised or Self-supervised Knowledge?} The proposed \pleakage is a practical reference to infer what kind of data we want to employ in NCD, images-only information or image-label pairs.
% Table \ref{tab:pt-leak-acc} indicates that \pleakage is consistent with various settings' accuracy.
% {E.g., in $L_2$-$U_1$, the \pleakage computed on the self-supervised model is larger than on the supervised one, which is consistent with accuracy, where $\mathbf{X}_u + \mathbf{X}_l$ outperforms $\mb{X}_u+ (\mb{X}_l, Y_l)$.}
In Table \ref{tab:pt-leak-acc} (PTL), we compute \pleakage via a supervised model and a self-supervised model based on pseudo labels. 
As suggested in Table \ref{tab:pt-leak-acc} (ACC), the \pleakage is consistent with the accuracy based on various datasets. 
For example, in $L_1$-$U_1$, the \pleakage computed on the supervised model is 
larger than the one computed in the self-supervised model, 
which is consistent with the accuracy, where the supervised method outperforms the self-supervised one.  
% Reversely, for $L_{2}$-$U_{1}$, $L_{1}$-$U_{2}$ and $L_{1.5}$-$U_1$, 
% the \pleakage computed on the self-supervised model is larger than the one computed in the supervised model, 
% which is again consistent with their relative performance.
% We omit the cases of $L_2$-$U_2$ and $L_{1.5}$-$U_1$, since their performance is within error margins.

% the \pleakage computed on the self-supervised model is larger than the supervised, 
% which is consistent with accuracy, where the method with $\mathbf{X}_u + \mathbf{X}_l$ 
% outperforms the one with $\mb{X}_u+ (\mb{X}_l, Y_l)$.
%
\paragraph{Combining Supervised and Self-supervised Knowledge} 
Instead of using either supervised knowledge or self-supervised knowledge from the labeled set,
we propose an effective and straightforward method, 
which smoothly combines both of them. 
We combine the labeled set's ground truth labels $y_{l_{GT}}$ with self-supervised pseudo labels $y_{l_{PL}}$.
$\alpha y_{l_{GT}} + (1 - \alpha) y_{l_{PL}}$ is the overall classification objective, where $\alpha \in [0, 1]$ is the supervised component weight.
This strategy has the same aim as UNO~\cite{fini2021unified} for $\alpha = 1$, but uses self-supervised pretraining instead of supervised.
As shown in Figure \ref{fig:alpha vs acc}, our proposed method improves CIFAR100 and ImageNet by 3\% and 5\%, respectively, compared to UNO.
Our method delivers significant improvements for low semantic similarity cases and competitive performances for high similarity cases.

\begin{table}[H]
    \centering
    \fontsize{9.3pt}{9.3pt}\selectfont
    \caption{Results showing the link between \pleakage (PTL) and accuracy on novel classes (ACC). 
    The \pleakage is computed based either on a supervised (SL) or self-supervised model (SSL), using ResNet18 in both cases. 
    The accuracy is obtained using the standard NCD setting ($\mb{X}_u+ (\mb{X}_l, Y_l)$) 
    for supervised learning, and self-supervised NCD ($\mb{X}_u+ \mb{X}_l$) for the self-supervised model.}
    \label{tab:pt-leak-acc}
    \begin{tabular}{@{}lccccccc@{}}
    \toprule
        & & \multicolumn{2}{c}{High similarity}   & \multicolumn{2}{c}{Medium similarity} & \multicolumn{2}{c}{Low similarity} \\ 
        \cmidrule(r){3-4} \cmidrule(r){5-6} \cmidrule(r){7-8}
   &  \multirow{-2}{*}{Model}  &
            $L_{1}-U_{1}$ &
            $L_2-U_2$ &
            $L_{1.5}-U_1$ &
            $L_{1.5}-U_2$ &
            $L_2-U_1$ &
            $L_1-U_2$ \\ 
            \midrule
            \multirow{2}{*}{PTL} & SSL &
            0.96 $\pm$ 0.01 &
            0.96 $\pm$ 0.02 &
            \textbf{1.14 $\pm$ 0.02} &
        \textbf{ 1.19 $\pm$ 0.01} &
            \textbf{1.05 $\pm$ 0.03} &
            \textbf{1.25 $\pm$ 0.03} \\ 
    & SL &
            \textbf{1.21 $\pm$ 0.02} &
        \textbf{ 1.21 $\pm$ 0.01} &
            1.03 $\pm$ 0.02 &
            0.98 $\pm$ 0.03 &
            0.99 $\pm$ 0.02 &
            0.96 $\pm$ 0.01\\
            \midrule
            \multirow{2}{*}{ACC} & SSL & 79.6 $\pm$ 1.1 &	89.2 $\pm$ 0.3 &	79.7 $\pm$ 1.0 &	\textbf{85.2 $\pm$ 1.0} &	\textbf{80.3 $\pm$ 0.3} &	\textbf{85.3 $\pm$ 0.5}
            \\
            & SL & \textbf{83.9 $\pm$ 0.6} &	88.3 $\pm$ 0.5 &	81.0 $\pm$ 0.6 &	82.0 $\pm$ 1.6 &	77.2 $\pm$ 0.8 &	77.5 $\pm$ 0.7\\
            \bottomrule
    \end{tabular}
    \vspace{-0.5cm}
    \end{table}

\section{Conclusion} 

We first offer a comprehensive ImageNet-based benchmark with varying levels of semantic similarity and show that semantic similarity affects NCD performance.
Second, we present \leakage, a semantic similarity metric.
Furthermore, we find that in low semantic similarity situations, labeled information may lead to inferior performance.
We propose two practical applications based on these findings: (i) Using \leakage to determine what data to use.
(ii) a straightforward approach that improves CIFAR100 and ImageNet by 3-5\%.

%\newpage

{\small
\bibliographystyle{splncs04nat}
\bibliography{nips2022}
}
\clearpage

\appendix
\section{Details of Section \ref{sec:higher greater}}
\label{sec:detail-of-benchmark}
\subsection{Assumption}
\label{sec:assumption}
Existing benchmarks consider the difficulty of NCD in terms of the labeled set from two aspects:
(1) the number of categories, e.g. \cite{fini2021unified} propose a more challenging benchmark called CIFAR100-50 (i.e., 50/50 classes for unlabeled/labeled set), compared to the commonly used CIFAR100-20 (i.e., 20/80 classes for unlabeled/labeled set).
(2) The number of images in each category, \cite{chi2021meta} propose to use less images for each labeled's class.

However, in addition to the number of categories and images, another significant factor is the semantic similarity between the two sets.
As mentioned in \cite{chi2021meta}, NCD is theoretically solvable when labeled and unlabeled sets share high-level semantic features.
Based on this, we conduct a further investigation with the assumption that more similar labeled sets (when the number of categories and images are fixed) can lead to better performance.
Intuitively, according to \autoref{fig:splits}, despite the fact that the labeled (e.g., pineapple, strawberry) and unlabeled (e.g., pear, jackfruit) classes are disjoint, 
if they derive from the same superclass (i.e., fruit), they have a higher degree of semantic similarity.
Conversely, when labeled (e.g., rose, lotus) and unlabeled (e.g., pear, jackfruit) classes are derived from distinct superclasses (i.e., labeled classes from flower while unlabeled classes from fruit), they are further apart semantically.
Consequently, we construct various semantic similarity labeled/unlabeled settings based on a hierarchical class structure and evaluate our assumption on CIFAR100 and ImageNet.

\subsection{Benchmark }
\label{sec:benchmark}
Existing benchmarks in the field were created without regard to the semantic similarity between labeled and unlabeled set.
Most works follow the standard splits introduced in \cite{han2019learning}.
In CIFAR10~\cite{krizhevsky2009learning}, the labeled set is made up of the first five classes in alphabetical order, and the unlabeled set of the remaining five.
A similar approach was taken with the commonly used CIFAR100-20 and CIFAR100-50 benchmarks.
A benchmark based on ImageNet~\cite{deng2009imagenet} has one labeled set, with 882 classes and three unlabeled sets.
Each of these unlabeled sets contains 30 classes, which were randomly selected from the remaining non-labeled classes \cite{vinyals2016matching, hsu2017learning, hsu2019multi, han2019learning}.
% In addition, these benchmarks usually only define one labeled set, and one unlabeled set, so it is not possible to compare 
% different semantic similarity settings.

To address this limitation and allow for an evaluation of our assumptions, we propose a new benchmark based on ImageNet including three different semantic similarity levels (high, medium and low).
As mentioned in Section~\ref{sec:assumption}, we separate labeled and unlabeled classes by leveraging ImageNet's underlying hierarchy. 
While ImageNet is based on the WordNet hierarchy~\cite{miller1995wordnet}, it is not well-suited for this purpose as discussed by \citep{santurkar2020breeds}. 
To address these issues, they propose a modified hierarchy and define multiple hierarchical classification tasks based on it.
While originally defined to measure the impact of subpopulation shift, they can also be used to define NCD tasks. 
% To allow for a systematic empirical evaluation of our assumptions, we propose a new benchmark based on ImageNet that addresses these shortcomings.
% To separate the classes into labeled and unlabeled sets based on their similarity, it is necessary to leverage their underlying hierarchy.
% While ImageNet is based on the WordNet hierarchy~\cite{miller1995wordnet}, this hierarchy is not well-suited for this purpose, as discussed by \citet{santurkar2020breeds}. 
% To address these issues, they propose a modified hierarchy and define multiple classification tasks on this hierarchy with a balanced number of subclasses per superclass. While originally defined to measure the impact of subpopulation shift, these superclasses can also be used to define NCD tasks. 
% Moving forward, we construct our benchmark based on the ENTITY-30 task. 

Our proposed benchmark is based on the ENTITY-30 task, which contains 240 ImageNet classes in total, with 30 superclasses and 8 subclasses for each superclasses.
For example, as shown in \autoref{fig:splits}, we define three labeled sets $L_{1}$, $L_{2}$ and $L_{1.5}$ and two unlabeled sets $U_{1}$ and $U_{2}$.
The sets $L_{1}$ and $U_{1}$ are selected from the first 15 superclasses, with 6 subclasses of each superclass assigned to $L_{1}$ and the other 2 assigned to $U_{1}$. 
The sets $L_{2}$ and $U_{2}$ are created from the second 15 superclasses in a similar fashion. 
Finally, $L_{1.5}$ is created by taking half the classes from $L_{1}$ and half of the classes from $L_{2}$.
Therefore, ($U_{1}$, $L_{1}$)/($U_{2}$, $L_{2}$) are highly related semantically, ($U_{1}$, $L_{2}$)/($U_{2}$, $L_{1}$) belong to the low semantic cases and ($U_{1}$, $L_{1.5}$)/($U_{2}$, $L_{1.5}$) are the medium cases.
Additionally, we also create four data settings on CIFAR100, with two high semantic cases and two low semantic cases by leveraging CIFAR100 hierarchical class structure.
Each case has 40 labeled classes and 10 unlabeled classes.
A full list of the labeled and unlabeled sets with their respective superclasses and subclasses can be found in Appendix \ref{sec:appendix-benchmark-splits}.

This benchmark setup allows us to systematically investigate the influence of the labeled set on a large benchmark dataset.
% Clearly, labeled sets and unlabeled sets that originate from the same superclasses are closely related, whereas sets from different superclasses are less related.
By keeping the unlabeled set constant and varying the used labeled set, we can isolate the influence of semantic similarity on NCD performance.

% \begin{figure}[H]
%     % \vspace{-0.0cm}
%     \centering
%     \includegraphics[width=1.0\linewidth]{figures/benchmark.pdf}
%     \vspace{-1.95cm}
%     %\setlength{\tabcolsep}{2.8pt} % Default value: 6pt
%     % \fontsize{9.1pt}{9.1pt}\selectfont
%     \caption{Illustration of how we construct the benchmark with varying levels of semantic similarity.
%     Unlabeled set $U_{1}$ and labeled set $L_{1}$ are from the same superclass (fruit), whereas unlabeled set $U_{2}$ and labeled set $L_{2}$ belong to another superclass (flower).
%     Labeled set $L_{1.5}$ is composed of half of $L_{1}$ and half of $L_{2}$.
%     If both the labeled and unlabeled classes are derived from the same superclass, i.e. ($U_{1}$, $L_{1}$) and ($U_{2}$, $L_{2}$), we consider them a \textbf{high semantic similarity} split.
%     In contrast, ($U_{1}$, $L_{2}$) and ($U_{2}$, $L_{1}$) are \textbf{low semantic similarity} splits, since the labeled and unlabeled classes are derived from distinct superclasses.
%     In addition, we consider ($U_{1}$, $L_{1.5}$) and ($U_{2}$, $L_{1.5}$) to have \textbf{medium semantic similarity} because half of $L_{1.5}$ share the same superclass as $U_{1}$.}
%     \vspace{-3mm}
%     \label{fig:splits}
%     \end{figure}

\subsection{Experimental Setup and Results}
\label{sec:3.3}

% \paragraph{Baselines.} 
To verify our assumption, we conduct experiments on four competitive baselines, including K-means \cite{macqueen1967some}, DTC \cite{han2019learning}, RS \cite{han2021autonovel},  {NCL} \cite{zhong2021neighborhood}   and UNO \cite{fini2021unified}.
We follow the baselines regarding hyperparameters and implementation details.
\paragraph{Results on CIFAR100}
In \autoref{tab:cifar100_benchmark}, $U_{1}$ and $U_{2}$ represent the unlabeled sets, while $L_{1}$ and $L_{2}$ represent the labeled sets.
$U_{1}$/$U_{2}$ and $L_{1}$/$L_{2}$ share the same super classes, while $U_{1}$/$U_{2}$ and $L_{2}$/$L_{1}$ belong to different super classes.
We evaluate 4 different labeled/unlabeled settings in CIFAR100, with 2 high semantic cases (i.e., ($U_{1}$, $L_{1}$) and ($U_{2}$, $L_{2}$)) and low semantic cases (i.e., ($U_{1}$, $L_{2}$) and ($U_{2}$, $L_{1}$)).
The gap between the high-similarity and the low-similarity settings is larger than 20\% for K-means, and reaches up to 12\% for more advanced methods.
The strong results of UNO across all splits show that a more difficult benchmark is needed to obtain clear results for future methods.

% the high related labeled set achieves higher clustering accuracy and higher normalized mutual information compared to the low related labeled set under various baselines.

\paragraph{Results on our proposed benchmark}
Similarly, in \autoref{tab:imagenet_benchmark}, the most similar labeled set generally obtains the best performance, followed by the medium and the least similar one.
Under the unlabeled set $U_{1}$, $L_{1}$ achieves the highest accuracy, with around 2-17\% improvement compared to $L_{2}$, and around 2-11\% improvement compared to $L_{1.5}$.
For the unlabeled set $U_{2}$, $L_{2}$ is the most similar set and obtains 8-14\% improvement compared to $L_{1}$, and around 5-14\% improvement compared to $L_{1.5}$.

\begin{table}[H]
\vspace{-2mm}
  \centering
    % \fontsize{9.5pt}{9.5pt}\selectfont
    \setlength{\tabcolsep}{9pt}
    % \captionsetup{width=0.96\linewidth} 
      \caption{ 
      Comparison of different combinations of labeled sets and unlabeled sets consisting of subsets of CIFAR100.
      The unlabeled set are denoted $U_{1}$ and $U_{2}$, while the labeled sets are called $L_{1}$ and $L_{2}$.
      $U_{1}$ and $L_{1}$ share the same set of superclasses, similar for $U_{2}$ and $L_{2}$.
      Thus, the pairs ($U_{1}$, $L_{1}$) and ($U_{2}$, $L_{2}$) are close semantically, but ($U_{1}$, $L_{2}$) and ($U_{2}$, $L_{1}$) are far apart.
      For all splits we report the mean and standard deviation of the clustering accuracy across multiple NCD methods.
      }
      \vspace{2mm}
      \label{tab:cifar100_benchmark}

      \begin{tabular}{@{}lcccc@{}}
      \toprule
      \multirow{2}{*}{Methods}        & \multicolumn{2}{c}{Unlabeled set $U_{1}$} & \multicolumn{2}{c}{Unlabeled set $U_{2}$} \\ 
                               \cmidrule(l){2-3}   \cmidrule(l){4-5} 
                               & $L_{1}$ - high     & $L_{2}$ - low     & $L_{1}$ - low     & $L_{2}$ - high    \\ 
                              \midrule
      K-means \cite{macqueen1967some}                & \textbf{61.0 $\pm$ 1.1}             & 37.7 $\pm$ 0.6           & 33.9 $\pm$ 0.5           & \textbf{55.4 $\pm$ 0.6}             \\
      DTC \cite{han2019learning}                     & \textbf{64.9 $\pm$ 0.3}             & 62.1 $\pm$ 0.3           & 53.6 $\pm$ 0.3           & \textbf{66.5 $\pm$ 0.4}             \\
      RS \cite{han2020automatically}                 & \textbf{78.3 $\pm$ 0.5}             & 73.7 $\pm$ 1.4           & 74.9 $\pm$ 0.5           & \textbf{77.9 $\pm$ 2.8}             \\
       {NCL} \cite{zhong2021neighborhood}               &  {\textbf{85.0 $\pm$ 0.6}}             &  {83.0 $\pm$ 0.3}           &  {72.5 $\pm$ 1.6}           &  {\textbf{85.6 $\pm$ 0.3}}             \\
      UNO \cite{fini2021unified}                     & \textbf{92.5 $\pm$ 0.2}             & 91.3 $\pm$ 0.8           & 90.5 $\pm$ 0.7           & \textbf{91.7 $\pm$ 2.2}             \\ 
      \bottomrule
      \end{tabular}
      \vspace{-5mm}
\end{table}

\section{Details of Section \ref{sec:transfer_leakage}}
\label{sec:detail-of-tleak}
\subsection{The Detailed Derivation Process of Transfer Leakage}
\label{lab:transfer leakage}
To summarize, \leakage measures the overall discrepancy of $\mb{p}(\mb{X}_u)$ under different new classes of the unlabeled measure $\mathbb{Q}$, which indicates the informative leakage from $\mathbb{P}$ to $\mathbb{Q}$. 
\begin{definition}[Transfer leakage]
	\label{def:T-leak}
	The \leakage of NCD prediction under $\mathbb{Q}$ based on the labeled conditional probability $\mathbb{P}_{Y_l|\mb{X}_l}$ is
	\begin{equation}
	\text{T-Leak}( \mathbb{Q}, \mathbb{P}) = \mathbb{E}_{\mathbb{Q}} \Big( \MMD_{\mathcal{H}}^2 \big( \mathbb{Q}_{ \mb{p}(\mb{X}_u) | Y_u }, \mathbb{Q}_{\mb{p}(\mb{X}_u') \mid Y_u'} \big) \Big),
	\end{equation}
	where $(\mb{X}_u, Y_u), (\mb{X}_u', Y_u') \sim \mathbb{Q}$ are independent copies, the expectation $\mathbb{E}_{\mathbb{Q}}$ is taken with respect to $Y_u$ and $Y_u'$ under $\mathbb{Q}$, and $\mb{p}(\mb{x})$ is the conditional probability under $\mathbb{P}_{Y_l|\mb{X}_l}$ on an unlabeled data $\mb{X}_u = \mb{x}$, which is defined as
	$
	\mb{p}(\mb{x}) = \big(\mathbb{P}\big( Y_l=c \mid \mb{X}_l = \mb{x} \big) \big)^{\intercal}_{c \in \mathcal{C}_l}.
	$
\end{definition}

Next, we give a finite sample estimate of \leakage. To proceed, we first rewrite \leakage as follows.
\begin{equation}
\text{T-Leak}( \mathbb{Q}, \mathbb{P}) = \sum_{c,c' \in \mathcal{C}_u; c \neq c'} \mathbb{Q}( Y_u = c, Y_u' = c') \MMD_{\mathcal{H}}^2 \big( \mathbb{Q}_{ \mb{p}(\mb{X}_u) | Y_u = c }, \mathbb{Q}_{\mb{p}(\mb{X}_u') \mid Y_u' = c'} \big),
\end{equation}
where the equality follows from the fact that $\MMD_{\mathcal{H}}^2 \big( \mathbb{Q}_{ \mb{p}(\mb{X}_u) | Y_u = c }, \mathbb{Q}_{\mb{p}(\mb{X}_u') \mid Y_u' = c} \big) = 0$.

  {Given an estimated probability $\widehat{\mathbb{P}}_{Y_l|\mb{X}_l}$ and an evaluation dataset $(\mb{x}_{u,i}, y_{u,i})_{i=1}^m$ under $\mathbb{Q}$}, we assess $\mb{x}_{u,i}$ on $\widehat{\mathbb{P}}_{Y_l | \mb{X}_l}$ as $\widehat{\mb p}(\mb{x}_{u,i}) = \big(\widehat{\mathbb{P}} \big( Y_l=c | \mb{X}_l = \mb{x}_{u,i} \big) \big)^{\intercal}_{c \in \mathcal{C}_u}$, then the empirical \leakage is computed as:
\begin{align}
	\label{eqn:emp_transfer_leak}
	\widehat{\text{T-Leak}}( \mathbb{Q}, \mathbb{P} ) &= \sum_{c, c' \in \mathcal{C}_u; c \neq c'} \frac{|\mathcal{I}_{u,c}| |\mathcal{I}_{u,c'}|}{m(m-1)} \widehat{\MMD}^2_{\mathcal{H}} \big( \mathbb{Q}_{ \widehat{\mb{p}}(\mb{X}_u) | Y_u = c }, \mathbb{Q}_{\widehat{\mb{p}}(\mb{X}_u') \mid Y_u' = c} \big),
\end{align}
where $\mathcal{I}_{u,c} = \{ 1 \leq i \leq m: y_{u,i} = c \}$ is the index set of unlabeled data with $y_{u,i} = c$, and $\widehat{\MMD}^2_{\mathcal{H}}$ is defined as:
\begin{align*}
    \widehat{\MMD}^2_{\mathcal{H}} ( \mathbb{Q}_{ \widehat{\mb{p}}(\mb{X}_u) | Y_u = c }, \mathbb{Q}_{\widehat{\mb{p}}(\mb{X}_u') \mid Y_u' = c} ) & = \frac{1}{ | \mathcal{I}_{u,c}| (| \mathcal{I}_{u,c}| - 1) } \sum_{i, j \in \mathcal{I}_{u,c}; i \neq j} K \big( \widehat{\mb{p}}(\mb{x}_{u,i}), \widehat{\mb{p}}(\mb{x}_{u,j}) \big) \\
    & + \frac{1}{|\mathcal{I}_{u,c'}|(|\mathcal{I}_{u,c'}| - 1)} \sum_{i, j \in \mathcal{I}_{u,c'}; i \neq j}  K \big( \widehat{\mb{p}}(\mb{x}_{u,i}), \widehat{\mb{p}}(\mb{x}_{u,j}) \big) \\ 
    & - \frac{2}{|\mathcal{I}_{u,c}| |\mathcal{I}_{u,c'}|} \sum_{i \in \mathcal{I}_{u,c}} \sum_{j \in \mathcal{I}_{u,c'}} K \big( \widehat{\mb{p}}(\mb{x}_{u,i}), \widehat{\mb{p}}(\mb{x}_{u,j}) \big).
\end{align*}
% \begin{align}
% 	\label{eqn:emp_transfer_leak}
% 	\widehat{\text{T-Leak}}( \mathbb{Q}, \mathbb{P} ) & = \frac{2}{m(m-1)} \sum_{1\leq i< i' \leq m}  \widehat{\MMD}^2_{\mathcal{H}} \big( \widehat{\mb p}( \mathcal{U}_{Y_{u,i}} ),  \widehat{\mb p}( \mathcal{U}_{ Y_{u,i'} } ) \big) \nonumber \\
% 	& = \sum_{c, c' \in \mathcal{C}^u; c \neq c'} \frac{|\mathcal{U}_c| |\mathcal{U}_{c'}|}{m(m-1)} \widehat{\MMD}^2_{\mathcal{H}} \big( \widehat{\mb{p}}(\mathcal{U}_c), \widehat{\mb{p}}(\mathcal{U}_{c'}) \big),
% \end{align}
\subsection{{Pseudo Transfer Leakage}}
{In practice, the ground-truth labels of the unlabeled set are difficult to acquire.
Therefore, we utilize pseudo labels derived from clustering algorithms, e.g., $k$-means, rather than true labels in computing transfer leakage,
and we named it as \pleakage}
\label{sec:p_tl}
\begin{definition}[Pseudo Transfer Leakage]
    \label{def:p_tl}
    \begin{equation}
   \widehat{\text{PT-Leak}}( \mathbb{Q}, \mathbb{P} ) = \sum_{c, c' \in \mathcal{C}_u; c \neq c'} \frac{| \widetilde{\mathcal{I}}_{u,c}| | \widetilde{\mathcal{I}}_{u,c'}|}{m(m-1)} \widehat{\text{MMD}}^2_{\mathcal{H}} \big( \mathbb{Q}_{ \widehat{\mathbf{p}}(\mathbf{X}_u) | \widetilde{Y}_u = c }, \mathbb{Q}_{\widehat{\mathbf{p}}(\mathbf{X}_u') \mid \widetilde{Y}_u' = c} \big)
    \end{equation}
    where $\widetilde{\mathcal{I}}_{u,c} = \{ 1 \leq i \leq m: \widetilde{y}_{u,i} = c \}$ is the index set of unlabeled data with $\widetilde{y}_{u,i} = c$, $\widetilde{y}_{u,i}$ is provided based on k-means on their representations $\widehat{\mathbf{s}}(\mathbf{x}_{u,i})$, and $\widehat{\mathbf{s}}(\mathbf{x}_{u,i})$ is the representation estimated from a supervised model or a self-supervised model. 
    $\widehat{\text{MMD}}^2_{\mathcal{H}}$ is defined similarly as in Section \ref{sec:transfer_leakage}. 
\end{definition}

\Pleakage, a practical evaluation of the similarity between the classified and unlabeled sets, could be a practical reference on whether to images-only or image-label pairs information from the labeled set.
% \Pleakage and \leakage differ in two aspects. 
% First, transfer leakage is computed based on the pretrained conditional probability $\mathbf{p}(\mathbf{x})$, 
% yet the pseudo transfer leakage is computed based on the pretrained representation $\widehat{\mathbf{s}}(\mathbf{x})$. 
% Second, 

\subsection{The Lower and Upper Bounds of Transfer Leakage}
Lemma \ref{lemma:tl-zero} shows the lower and upper bounds of \leakage, and provide a theoretical justification that it is a effective quantity to measure the similarity between labeled and unlabeled datasets. % For multi-class discrepancy, we mimic the one-vs-rest nature by taking the minimum discrepancy over all rest classes, which takes the discrepancy to the closest class. 
\begin{lemma}
	\label{lemma:tl-zero}
	  {Let $\kappa := \max_{c \in \mathcal{C}_u} \mathbb{E}_{\mathbb{Q}}\big( \sqrt{K ( \mb{p}(\mb{X}_u), \mb{p}(\mb{X}_u) )} | Y_u =c \big)  < \infty$, then $0 \leq \text{T-Leak}( \mathbb{Q}, \mathbb{P} ) \leq 4 \kappa^2.$}
	Moreover, $\text{T-Leak}( \mathbb{Q}, \mathbb{P} ) = 0$ if and only if $Y_u$ is independent with $\mb{p}(\mb{X}_u)$, that is, for any $c \in \mathcal{C}_u$:
	\begin{equation}
		\label{eqn:tl-indep}
		\mathbb{Q}\big( Y_u = c \mid \mb{p}(\mb{X}_u) \big) = \mathbb{Q}(Y_u = c),
	\end{equation}
	yielding that $\mb{p}(\mb{X}_u)$ is useless in NCD on $\mathbb{Q}$. 
\end{lemma}
Note that $\kappa$ can be explicitly computed for many common used kernels, for example, $\kappa = 1$ for a Gaussian or Laplacian kernel. 
From Lemma \ref{lemma:tl-zero}, $\text{T-Leak}( \mathbb{Q}, \mathbb{P} ) = 0$ is equivalent to $Y_u$ is independent with $\mb{p}(\mb{X}_u)$, 
which matches our intuition of no leakage. Alternatively, if $Y_u$ is dependent with $\mb{p}(\mb{X}_u)$, 
we justifiably believe that the information of $Y_l|\mb{X}_l$ can be used to facilitate NCD, 
Lemma \ref{lemma:tl-zero} tells that $\text{T-Leak}( \mathbb{Q}, \mathbb{P} ) > 0$ in this case. 
Therefore, Lemma \ref{lemma:tl-zero} reasonably suggests that the proposed \leakage 
is an effective metric to detect if the labeling information in $\mathbb{P}$ is useful to NCD on $\mathbb{Q}$.

\noindent \textbf{Proof of Lemma 1.} We first show the upper bound of the transfer leakage. According to Lemma 3 in \cite{gretton2012kernel}, we have
    \begin{align*}
        \text{T-Leak}(\mathbb{Q}, \mathbb{P}) & = \mathbb{E}_{\mathbb{Q}} \big( \MMD^2( \mathbb{Q}_{\mb{s}(\mb{X}_u)|Y_u}, \mathbb{Q}_{\mb{s}(\mb{X}_u')|Y_u'} ) \big) = \mathbb{E}_{\mathbb{Q}} \Big( \big\| \mu_{\mathbb{Q}_{\mb{s}(\mb{X}_u)|Y_u}} - \mu_{\mathbb{Q}_{\mb{s}(\mb{X}_u')|Y_u'}} \big\|_{\mathcal{H}}^2 \Big) \\
        & \leq \max_{c,c' \in \mathcal{C}_u} \big\| \mu_{\mathbb{Q}_{\mb{s}(\mb{X}_u)|Y_u=c}} - \mu_{\mathbb{Q}_{\mb{s}(\mb{X}_u')|Y_u'=c'}} \big\|_{\mathcal{H}}^2 \leq 4 \max_{c \in \mathcal{C}_u} \| \mu_{\mathbb{Q}_{\mb{s}(\mb{X}_u)|Y_u=c}} \|^2_{\mathcal{H}} \\
        & = 4 \max_{c \in \mathcal{C}_u} \langle  \mathbb{E}_{\mathbb{Q}} \big(K( \mb{s}(\mb{X}_u), \cdot) | Y_u = c \big) , \mathbb{E}_{\mathbb{Q}} \big(  K(\mb{s}(\mb{X}_u'), \cdot) | Y_u' = c \big) \rangle_{\mathcal{H}} \\
        & = 4 \max_{c \in \mathcal{C}_u} \mathbb{E}_{\mathbb{Q}} \big( \langle K( \mb{s}(\mb{X}_u), \cdot) , K(\mb{s}(\mb{X}_u'), \cdot) \rangle_{\mathcal{H}} | Y_u = c, Y_u'=c \big) \\
        & \leq 4 \max_{c \in \mathcal{C}_u} \mathbb{E}_{\mathbb{Q}} \Big( \big\| K(\mb{s}(\mb{X}_u), \cdot) \big\|_{\mathcal{H}}  \big\| K(\mb{s}(\mb{X}_u'), \cdot) \big\|_{\mathcal{H}} | Y_u = c, Y_u'=c \Big) \\
        & = 4 \max_{c \in \mathcal{C}_u} \mathbb{E}_{\mathbb{Q}} \big( \sqrt{K(\mb{s}(\mb{X}_u), \mb{s}(\mb{X}_u))} | Y_u = c \big) \mathbb{E}_{\mathbb{Q}} \big( \sqrt{K(\mb{s}(\mb{X}_u'), \mb{s}(\mb{X}_u'))} | Y_u' = c \big) \leq 4 \kappa^2,
    \end{align*}
where $\mu_{\mathbb{Q}_{\mb{s}(\mb{X}_u)|Y_u}} := \mathbb{E}_{\mathbb{Q}} \big( K(\mb{s}(\mb{X}_u), \cdot) | Y_u \big)$ is the kernel mean embedding of the measure $\mathbb{Q}_{\mb{s}(\mb{X}_u)|Y_u}$ \cite{gretton2012kernel}, the second inequality follows from the triangle inequality in the Hilbert space, the fourth equality follows from the fact that $\mathbb{E}_{\mathbb{Q}}$ is a linear operator, the second last inequality follows from the Cauchy-Schwarz inequality, and the last equality follows the reproducing property of $K(\cdot, \cdot)$.

Next, we show the if and only if condition for $\text{T-Leak}(\mathbb{Q}, \mathbb{P}) = 0$. Assume that $\mathbb{Q}(Y_u=c) > 0$ for all $c \in \mathcal{C}_u$. According to Theorem 5 in \cite{gretton2012kernel}, we have
\begin{align*}
    \text{T-Leak}(\mathbb{Q}, \mathbb{P}) = 0 \quad  \iff \quad \mathbb{Q}\big( \mb{s}(\mb{x}) | Y_u = c \big) = q_0(\mb{x}), \text{ for } c \in \mathcal{C}_u, \mb{x} \in \mathcal{X}_u.
\end{align*}
Note that 
\begin{align*}
    1 = \sum_{c \in \mathcal{C}_u} \mathbb{Q}( Y_u=c | \mb{s}(\mb{x}) ) = \sum_{c \in \mathcal{C}_u} \frac{\mathbb{Q}( \mb{s}(\mb{x}) | Y_u = c ) \mathbb{Q}(Y_u=c) }{ \mathbb{Q}(\mb{s}(\mb{x})) } = \sum_{c \in \mathcal{C}_u} \frac{ q_0(\mb{x}) \mathbb{Q}(Y_u=c) }{ \mathbb{Q}(\mb{s}(\mb{x})) } = \frac{ q_0(\mb{x}) }{ \mathbb{Q}(\mb{s}(\mb{x})) },
\end{align*}
yielding that $\mathbb{Q}\big(\mb{s}(\mb{x})|Y_u=c\big) = \mathbb{Q}( \mb{s}(\mb{x}) )$, for $c \in \mathcal{C}_u, \mb{x} \in \mathcal{X}_u$. This is equivalent to,
\begin{align*}
    \mathbb{Q}\big( Y_u=c | \mb{s}(\mb{x}) \big) = \frac{ \mathbb{Q}\big( \mb{s}(\mb{x}) | Y_u=c \big) \mathbb{Q}(Y_u=c) }{ \mathbb{Q}(\mb{s}(\mb{x})) } = \mathbb{Q}( Y_u=c ).
\end{align*}
This completes the proof. \EOP

\begin{figure*}
  \centering
  \begin{tikzpicture}%[trim axis left,trim axis right]
  \centering
    \begin{groupplot}[
        group style={group size= 2 by 1, 
                    horizontal sep=0.1\textwidth},
        width=\textwidth,
        ]
        \nextgroupplot[
          axis x line=bottom,
          axis y line=left,
          clip=false,
          width={0.47\textwidth},
          xtick={0.3, 0.5, 0.7},
          ytick={30, 40, 50, 60, 70, 80, 90},
          y tick label style={font=\small},
          x tick label style={font=\small}, 
          xlabel={Transfer-leakage},
          ylabel={Accuracy in \%},
          legend to name={CommonLegend},
          legend columns = 5,
          legend style={anchor=south,fill=none, draw=none},
          xmin=0.2,
          xmax=0.8,
          ymin=20.0,
          ymax=93.0,
          u1/.style={mark=triangle*, mark size=3pt, dashed},
          u2/.style={mark=square*},
          u1 legend/.style={
          legend image code/.code={
              \draw [/pgfplots/u1] plot coordinates {(0cm,0cm)};
              }
          },
          uno/.style={Magenta},
          ncl/.style={NavyBlue},
          kmeans/.style={OliveGreen},
          ]
          \addplot [forget plot, u1, uno, mark=none] table { 
              0.71    83.89
              0.53    80.95
              0.35    77.19
              };
          \addplot [forget plot, u1, ncl, mark=none] table { 
              0.71    75.10
              0.53    74.34
              0.35    71.58
              };
          \addplot [forget plot, u1, kmeans, mark=none] table { 
              0.71    41.11
              0.53    30.24
              0.35    23.27
              };
  
          \addplot [forget plot, u2, uno, mark=none] table {
              0.33    77.45
              0.5     82.02
              0.72    88.35
              };
          \addplot [forget plot, u2, ncl, mark=none] table { 
              0.33    61.27
              0.5     70.49
              0.72    75.09
              };
          \addplot [forget plot, u2, kmeans, mark=none] table { 
              0.33    21.19
              0.5     29.78
              0.72    44.96
              };

          \addplot [forget plot, only marks, mark=square*, uno] table {
              0.72    88.35
              };
          \addplot [forget plot, only marks, mark=square*, uno] table {
              0.71    83.89
              };
          \addplot [forget plot, only marks, mark=triangle*, mark size=3pt, uno] table {
              0.5     82.02
              };
          \addplot [forget plot, only marks, mark=triangle*, mark size=3pt, uno] table {
              0.53    80.95
              };
          \addplot [forget plot, only marks, mark=*, uno] table {
              0.33    77.45
              };
          \addplot [forget plot, only marks, mark=*, uno] table {
              0.35    77.19
              };

          \addplot [forget plot, only marks, mark=square*, ncl] table {
              0.72    75.09
              };
          \addplot [forget plot, only marks, mark=square*, ncl] table {
              0.71    75.10
              };
          \addplot [forget plot, only marks, mark=triangle*, mark size=3pt, ncl] table {
              0.5     70.49
              };
          \addplot [forget plot, only marks, mark=triangle*, mark size=3pt, ncl] table {
              0.53    74.34
              };
          \addplot [forget plot, only marks, mark=*, ncl] table {
              0.33    61.27
              };
          \addplot [forget plot, only marks, mark=*, ncl] table {
              0.35    71.58
              };

          \addplot [forget plot, only marks, mark=square*, kmeans] table {
              0.72    44.96
              };
          \addplot [forget plot, only marks, mark=square*, kmeans] table {
              0.71    41.11
              };
          \addplot [forget plot, only marks, mark=triangle*, mark size=3pt, kmeans] table {
              0.5     29.78
              };
          \addplot [forget plot, only marks, mark=triangle*, mark size=3pt, kmeans] table {
              0.53    30.24
              };
          \addplot [forget plot, only marks, mark=*, kmeans] table {
              0.33    21.19
              };
          \addplot [forget plot, only marks, mark=*, kmeans] table {
              0.35    23.27
              };

          \addlegendimage{mark=*}
          \addlegendentry{Low similarity}
          \addlegendimage{mark=triangle*, mark size=3pt}
          \addlegendentry{Medium similarity}
          \addlegendimage{mark=square*}
          \addlegendentry{High similarity}
          \addlegendimage{empty legend}
          \addlegendentry{}
          \addlegendimage{empty legend}
          \addlegendentry{}
          
          \addlegendimage{mark=square*, Magenta}
          \addlegendentry{\small UNO}
          \addlegendimage{mark=square*, NavyBlue}
          \addlegendentry{\small NCL}
          \addlegendimage{mark=square*, OliveGreen}
          \addlegendentry{\small K-means}
          \addlegendimage{mark=none, dashed}
          \addlegendentry{\small $U_{1}$}
          \addlegendimage{mark=none}
          \addlegendimage{mark=none}
          \addlegendentry{\small $U_{2}$}

          \addplot [forget plot, mark=none, gray, dashed, opacity=0.5] coordinates {(0.53, 20) (0.53, 90)};
          \addplot [forget plot, mark=none, gray, opacity=0.5] coordinates {(0.5, 20) (0.5, 90)};
  
          \addplot [forget plot, mark=none, gray, dashed, opacity=0.5] coordinates {(0.71, 20) (0.71, 90)};
          \addplot [forget plot, mark=none, gray, opacity=0.5] coordinates {(0.72, 20) (0.72, 90)};
  
          \addplot [forget plot, mark=none, gray, dashed, opacity=0.5] coordinates {(0.35, 20) (0.35, 90)};
          \addplot [forget plot, mark=none, gray, opacity=0.5] coordinates {(0.33, 20) (0.33, 90)};
    \nextgroupplot[
          axis x line=bottom,
          axis y line=left,
          clip=false,
          width={0.47\textwidth},
          xtick={0.9, 1.1, 1.3},
          ytick={30, 40, 50, 60, 70, 80, 90},
          y tick label style={font=\small},
          x tick label style={font=\small}, 
          xlabel={Pseudo Transfer-leakage},
          ylabel={Accuracy in \%},
          legend columns = 3,
          legend style={at={(0.5,1.0)},anchor=south,fill=none, draw=none},
          xmin=0.85,
          xmax=1.35,
          ymin=20.0,
          ymax=93.0,
          u1/.style={mark=triangle*, mark size=3pt, dashed},
          u2/.style={mark=square*},
          u1 legend/.style={
          legend image code/.code={
              \draw [/pgfplots/u1] plot coordinates {(0cm,0cm)};
              }
          },
          uno/.style={Magenta},
          ncl/.style={NavyBlue},
          kmeans/.style={OliveGreen},
          ]
          \addplot [forget plot, u1, uno, mark=none] table { 
              1.21    83.89
              1.02    80.95
              0.99    77.19
              };
          \addplot [forget plot, u1, ncl, mark=none] table { 
              1.21    75.10
              1.02    74.34
              0.99    71.58
              };
          \addplot [forget plot, u1, kmeans, mark=none] table { 
              1.21    41.11
              1.02    30.24
              0.99    23.27
              };
  
          \addplot [forget plot, u2, uno, mark=none] table {
              0.96    77.45
              0.98    82.02
              1.21    88.35
              };
          \addplot [forget plot, u2, ncl, mark=none] table { 
              0.96    61.27
              0.98    70.49
              1.21    75.09
              };
          \addplot [forget plot, u2, kmeans, mark=none] table { 
              0.96    21.19
              0.98     29.78
              1.21    44.96
              };

          \addplot [forget plot, only marks, mark=square*, uno] table {
              1.21    88.35
              };
          \addplot [forget plot, only marks, mark=square*, uno] table {
              1.21    83.89
              };
          \addplot [forget plot, only marks, mark=triangle*, mark size=3pt, uno] table {
              0.98     82.02
              };
          \addplot [forget plot, only marks, mark=triangle*, mark size=3pt, uno] table {
              1.02    80.95
              };
          \addplot [forget plot, only marks, mark=*, uno] table {
              0.96    77.45
              };
          \addplot [forget plot, only marks, mark=*, uno] table {
              0.99    77.19
              };

          \addplot [forget plot, only marks, mark=square*, ncl] table {
              1.21    75.09
              };
          \addplot [forget plot, only marks, mark=square*, ncl] table {
              1.21    75.10
              };
          \addplot [forget plot, only marks, mark=triangle*, mark size=3pt, ncl] table {
              0.98     70.49
              };
          \addplot [forget plot, only marks, mark=triangle*, mark size=3pt, ncl] table {
              1.02    74.34
              };
          \addplot [forget plot, only marks, mark=*, ncl] table {
              0.96    61.27
              };
          \addplot [forget plot, only marks, mark=*, ncl] table {
              0.99    71.58
              };

          \addplot [forget plot, only marks, mark=square*, kmeans] table {
              1.21    44.96
              };
          \addplot [forget plot, only marks, mark=square*, kmeans] table {
              1.21    41.11
              };
          \addplot [forget plot, only marks, mark=triangle*, mark size=3pt, kmeans] table {
              0.98    29.78
              };
          \addplot [forget plot, only marks, mark=triangle*, mark size=3pt, kmeans] table {
              1.02    30.24
              };
          \addplot [forget plot, only marks, mark=*, kmeans] table {
              0.96    21.19
              };
          \addplot [forget plot, only marks, mark=*, kmeans] table {
              0.99    23.27
              };

          \addplot [forget plot, mark=none, gray, dashed, opacity=0.5] coordinates {(1.02, 20) (1.02, 90)};
          \addplot [forget plot, mark=none, gray, opacity=0.5] coordinates {(0.98, 20) (0.98, 90)};
  
          \addplot [forget plot, mark=none, gray, dashed, opacity=0.5] coordinates {(1.21, 20) (1.21, 90)};
          \addplot [forget plot, mark=none, gray, opacity=0.5] coordinates {(1.21, 20) (1.21, 90)};
  
          \addplot [forget plot, mark=none, gray, dashed, opacity=0.5] coordinates {(0.99, 20) (0.99, 90)};
          \addplot [forget plot, mark=none, gray, opacity=0.5] coordinates {(0.96, 20) (0.96, 90)};
  \end{groupplot}
  \path (group c1r1.north east) -- node[above]{\pgfplotslegendfromname{CommonLegend}} (group c2r1.north west);
  \end{tikzpicture}
  \caption{
  Experiments on \leakage and \pleakage.
  Each line stands for one unlabeled set from our proposed ImageNet-based benchmark, and each point on the line for one labeled / unlabeled split.
  On the left, we measure the \leakage and the clustering accuracy obtained using UNO, NCL and $k$-means for each split.
  As expected, there is a positive correlation between semantic similarity and \leakage as well as between \leakage and accuracy. 
  On the right, we replace \leakage with \pleakage obtained using $k$-means clustering. 
  The comparison shows that \pleakage can in practice be used as a proxy for \leakage.}
  \label{fig:leakage}
\end{figure*}
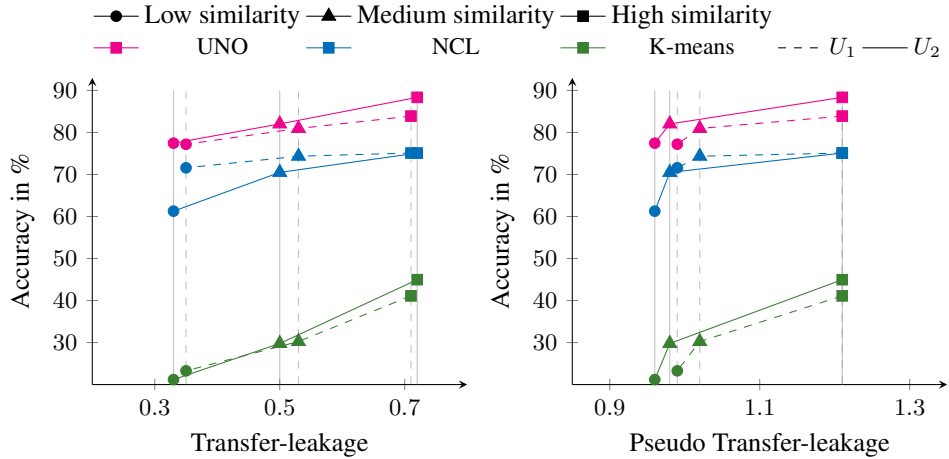

\subsection{Experiments and Results}
\label{sec:exp}
\paragraph{{Experimental Setup and Hyperparameters}}
To calculate the \leakage/\pleakage, we employ ResNet18~\citep{he2016deep} as the backbone for both datasets following \citep{han2019learning, han2021autonovel, fini2021unified}.
Known-class data and unknown-class data are selected based on semantic similarity, as mentioned in Section \ref{sec:higher greater}.
We first apply fully supervised learning to the labeled data for each data set to obtain the pretrained model.
Then, we feed the unlabeled data to the pretrained model to obtain its representation.
Lastly, we calculate the \leakage/\pleakage based on the pretrained model and the unlabeled samples' representation.
Speciallly, for \pleakage, we apply clustering methods to generate the pseudo labels.
For the first step, batch size is set to 512 for both datasets.
We use an SGD optimizer with momentum 0.9, and weight decay 1e-4.
The learning rate is governed by a cosine annealing learning rate schedule with a base learning rate of 0.1, a linear warmup of 10 epochs, and a minimum learning rate of 0.001. 
We pretrain the backbone for 200/100 epochs for CIFAR-100/ImageNet.

% \paragraph*{Hyperparameters}
% For the first step, batch size is set to 512 for both datasets
% We use an SGD optimizer with momentum 0.9, and weight decay 1e-4.
% The learning rate is governed by a cosine annealing learning rate schedule with a base learning rate of 0.1, a linear warmup of 10 epochs, and a minimum learning rate of 0.001. 
% We pretrain the backbone for 200/100 epochs for CIFAR-100/ImageNet.

\paragraph{Results}
\autoref{fig:leakage} shows semantic similarity, \leakage/\pleakage and NCD performance on ImageNet, in which the same color corresponds to the same unlabeled set.
As expected, splits that have a higher semantic similarity yield both higher \leakage and \pleakage.
Alternatively, there is a consistent positive correlation between \leakage/\pleakage and NCD accuracy, which confirms the validity of \leakage/\pleakage as a metric in quantifying semantic similarity and the difficulty of a particular NCD problem.

% \begin{table}[H]
%     \centering
%     \caption{
%     {Experiments on the stability of \leakage. 
%     To obtain the standard deviation we recompute the \leakage 10 times using bootstrap sampling. 
%     The results show that \leakage has a low random variation.}
%     }
%     \label{tab:transfer leakage std}
%     \begin{tabular}{@{}lllcc@{}}
    
%     \toprule 
%     Dataset & Unlabeled Set & Labeled Set & \Leakage \\
%     \midrule
%     \multirow{4}{*}{CIFAR100} & \multirow{2}{*}{$U_{1}$} & $L_{1}$ & 0.62 $\pm$ 0.01 \\
%      & & $L_{2}$ & 0.28 $\pm$ 0.01 \\
%      \cmidrule(l){3-4} 
%      & \multirow{2}{*}{$U_{2}$} & $L_{1}$ & 0.33 $\pm$ 0.01 \\
%      & & $L_{2}$ & 0.77 $\pm$ 0.02 \\

%      \cmidrule(l){2-4} 

%      \multirow{6}{*}{ImageNet} & \multirow{3}{*}{$U_{1}$} & $L_{1}$ & 0.71 $\pm$ 0.01 \\
%      & & $L_{1.5}$ & 0.54 $\pm$ 0.01 \\
%      & & $L_{2}$ & 0.36 $\pm$ 0.01 \\
%      \cmidrule(l){3-4} 
%      & \multirow{3}{*}{$U_{2}$} & $L_{1}$ & 0.33 $\pm$ 0.00 \\
%      & & $L_{1.5}$ & 0.50 $\pm$ 0.01 \\
%      & & $L_{2}$ & 0.72 $\pm$ 0.01 \\
%     \bottomrule
%     \end{tabular}
% \end{table}

\begin{table}[H]
    \centering
    \caption{
    {Experiments on the stability of \pleakage and \leakage. 
    To obtain the standard deviation we recompute the \leakage and \pleakage 10 times using bootstrap sampling. 
    The results show that both \leakage has a low random variation.}
    }
    \label{tab:pseudo transfer leakage std}
    \begin{tabular}{@{}lllccc@{}}
    
    \toprule 
    Dataset & Unlabeled Set & Labeled Set & \Leakage & {\Pleakage}\\
    \midrule
    \multirow{4}{*}{CIFAR100} & \multirow{2}{*}{$U_{1}$} & $L_{1}$ & 0.62 $\pm$ 0.01 & 0.89 $\pm$ 0.06\\
     & & $L_{2}$ & 0.28 $\pm$ 0.01 & 0.74 $\pm$ 0.04 \\
     \cmidrule(l){3-5} 
     & \multirow{2}{*}{$U_{2}$} & $L_{1}$ & 0.33 $\pm$ 0.01 & 0.73 $\pm$ 0.02 \\
     & & $L_{2}$ & 0.77 $\pm$ 0.02 & 1.24 $\pm$ 0.01\\

     \cmidrule(l){2-5} 

     \multirow{6}{*}{ImageNet} & \multirow{3}{*}{$U_{1}$} & $L_{1}$ & 0.71 $\pm$ 0.01 & 1.21 $\pm$ 0.02\\
     & & $L_{1.5}$ & 0.54 $\pm$ 0.01 & 1.03 $\pm$ 0.02\\
     & & $L_{2}$ & 0.36 $\pm$ 0.01 &  0.99 $\pm$ 0.02\\
     \cmidrule(l){3-5} 
     & \multirow{3}{*}{$U_{2}$} & $L_{1}$ & 0.33 $\pm$ 0.00 & 0.96$\pm$ 0.01\\
     & & $L_{1.5}$ & 0.50 $\pm$ 0.01  & 0.98$\pm$ 0.03\\
     & & $L_{2}$ & 0.72 $\pm$ 0.01  & 1.21$\pm$ 0.01\\
    \bottomrule
    \end{tabular}
\end{table}

\begin{table}[H]
    \centering
    \setlength{\tabcolsep}{2.5pt}
    \caption{Experiments on \pleakage under three clustering methods, i.e., $k$-means, GMM and agglomerative, and each setting is repeated for 10 times. }
    \label{tab:p-leak}
        % \vspace{-2cm}
    \begin{tabular}{@{}lcccccc@{}}
    \toprule
    \multirow{2}{*}{Method} & \multicolumn{3}{c}{Unlabeled set $U_{1}$}     & \multicolumn{3}{c}{Unlabeled set $U_{2}$}  \\
    \cmidrule(l){2-4} \cmidrule(l){5-7}
         & $L_{1}$ - high & $L_{1.5}$ - medium & $L_{2}$ - low & $L_{1}$ - low & $L_{1.5}$ - medium & $L_{2}$ - high\\
    % \multirow{2}{*}{Method} & \multicolumn{3}{c}{$U_1$} & \multicolumn{3}{c}{U2} \\  \cmidrule(l){2-4} \cmidrule(l){5-7}
    %  & L1    & L1.5    & L2   & L1    & L1.5    & L2   \\
    \midrule
    $k$-means       & 1.23 $\pm$ 0.03 & 1.02 $\pm$ 0.03 & 0.99 $\pm$ 0.02 & 0.96 $\pm$ 0.01 & 0.99 $\pm$ 0.03 & 1.24 $\pm$ 0.02 \\
    GMM           & 0.79 $\pm$ 0.01 & 0.69 $\pm$ 0.02 & 0.56 $\pm$ 0.02 & 0.58 $\pm$ 0.02 & 0.68 $\pm$ 0.04 & 0.91 $\pm$ 0.02 \\
    Agglomerative & 1.17 $\pm$ 0.00 & 0.96 $\pm$ 0.00 & 0.87 $\pm$ 0.00 & 0.83 $\pm$ 0.00 & 0.89 $\pm$ 0.00 & 1.15 $\pm$ 0.00 \\ \bottomrule
    \end{tabular}
        % \vspace{5cm}
    \end{table}

\section{Details of Section \ref*{sec:observation}}
\label{sec:detail-of-observation}

\subsection{Counterintuitive Observations}
The motivation behind NCD is that prior supervised knowledge from labeled data can help improve the clustering of the unlabeled set.
However, we have counterintuitive results:
supervised information from a labeled set may result in suboptimal outcomes as compared to using exclusively self-supervised knowledge.
To investigate, we conduct experiments in the following settings:

\vspace{2\baselineskip}

\begin{enumerate}[label={(\arabic*)}]
    \vspace{-0.9cm}
    \item Using the unlabeled set, $\mb{X}_u$
    \vspace{-0.1cm}
    \item Using the unlabeled set and the labeled set's images without labels, $\mb{X}_u+ \mb{X}_l$
    \vspace{-0.1cm}
    \item Using the unlabeled set and the whole labeled set, $\mb{X}_u+ (\mb{X}_l, Y_l)$, (i.e., standard NCD).
\end{enumerate}

Specifically, for (1), NCD is degenerated to unsupervised learning (i.e., clustering on $\mb{X}_u$).
For (2), even though we do not use the labels, we can still try to extract the knowledge of the labeled set via self-supervised learning.
By comparing (1) and (3), we can estimate the total performance gain caused by adding the labeled set.
The comparison between (1) and (2) as well as (2) and (3) allows us to further disentangle the performance gain according to the components of the labeled set.

% More details on these modifications as well as the used hyperparameters can be found in the Appendix.

\begin{table}[H]
\centering
\caption{Comparison of different data settings on CIFAR100 and the unlabeled set $U_{1}$ of our proposed benchmark.
We report the mean and the standard error of the clustering accuracy of UNO.
As UNO uses multiple unlabeled heads, we report their mean accuracy, as well as that of the head with the lowest loss.
Setting (1) uses only the unlabeled set, whereas (2) uses both the unlabeled set and the labeled set's images without labels.
Setting (3) represents the standard NCD setting, i.e., using the unlabeled set and the whole labeled set.
Counterintuitively, in CIFAR100-50 and on the low similarity case of our benchmark, we can achieve better performance without using the labeled set's labels.
}
\label{tab:counterintuitive}
\begin{tabular}{@{}llcccc@{}}

      \toprule
% \multicolumn{1}{l}{}       & \multicolumn{1}{l}{} &         & High semantic  & Middle semantic  & Low semantic  \\
% \multirow{2}{*}{UNO}           &            & CIFAR100-50       & ($L_{1}$-$U_{1}$)      &    ($L_{1.5}$-$U_{1}$)        &    ($L_{2}$-$U_{1}$)     \\
\multirow{2}{*}{UNO} & \multirow{2}{*}{Setting} 
 & \multirow{2}{*}{CIFAR100-50} & \multicolumn{3}{c}{ImageNet $U_{1}$} \\
 \cmidrule(l){4-6} 
 & & & $L_{1}$ - high & $L_{1.5}$ - medium & $L_{2}$ - low\\
\midrule
\multirow{3}{*}{Avg head}  & (1) $\mb{X}_u$                 & 54.2 $\pm$ 0.3          & 69.2 $\pm$ 0.7          & 69.2 $\pm$ 0.7          & 69.2 $\pm$ 0.7             \\
                          & (2) $\mb{X}_u+ \mb{X}_l$        & \textbf{63.4 $\pm$ 0.4}          & 74.9 $\pm$ 0.3          & 77.6 $\pm$ 0.9          & \textbf{77.9 $\pm$ 1.1}   \\
                          & (3) $\mb{X}_u+ (\mb{X}_l, Y_l)$         & 61.7 $\pm$ 0.3          & \textbf{81.7 $\pm$ 1.0} & \textbf{80.3 $\pm$ 0.4} & 74.6 $\pm$ 0.3             \\
                          \midrule
\multirow{3}{*}{Best head} & (1) $\mb{X}_u$                 & 54.9 $\pm$ 0.4          & 70.5 $\pm$ 1.2          & 70.5 $\pm$ 1.2          & 70.5 $\pm$ 1.2            \\
                          & (2) $\mb{X}_u+ \mb{X}_l$          & \textbf{64.1 $\pm$ 0.4} & 79.6 $\pm$ 1.1          & 79.7 $\pm$ 1.0          & \textbf{80.3 $\pm$ 0.3}   \\
                          & (3) $\mb{X}_u+ (\mb{X}_l, Y_l)$        & 62.2 $\pm$ 0.2          & \textbf{83.9 $\pm$ 0.6} & \textbf{81.0 $\pm$ 0.6} & 77.2 $\pm$ 0.8            \\
  \bottomrule
\end{tabular}
% \vspace{-1cm}
\end{table}

\begin{table}[H]
    \centering
    \caption{{Comparison of different data settings on the unlabeled set $U_{2}$ of our proposed benchmark, similar to Table \ref{tab:counterintuitive}. 
    We report the mean and the standard error of the clustering accuracy of UNO.}}
    % As UNO uses multiple unlabeled heads, we report their mean accuracy, as well as that of the head with the lowest loss.
    % Setting (1) uses only the unlabeled set, whereas (2) uses both the unlabeled set and the labeled set's images without labels.
    % Setting (3) represents the standard NCD setting, i.e., using the unlabeled set and the whole labeled set.}
    
    \label{tab:counterintuitive u2}
    \begin{tabular}{@{}llccc@{}}
    
          \toprule
    % \multicolumn{1}{l}{}       & \multicolumn{1}{l}{} &         & High semantic  & Middle semantic  & Low semantic  \\
    % \multirow{2}{*}{UNO}           &            & CIFAR100-50       & ($L_{1}$-$U_{1}$)      &    ($L_{1.5}$-$U_{1}$)        &    ($L_{2}$-$U_{1}$)     \\
    \multirow{2}{*}{UNO} & \multirow{2}{*}{Setting} 
     & \multicolumn{3}{c}{ImageNet $U_{2}$} \\
     \cmidrule(l){3-5} 
     & & $L_{1}$ - low & $L_{1.5}$ - medium & $L_{2}$ - high\\
    \midrule
    \multirow{3}{*}{Avg head}  & (1) $\mb{X}_u$                 & 68.4 $\pm$ 0.6          & 68.4 $\pm$ 0.6          & 68.4 $\pm$ 0.6             \\
                              & (2) $\mb{X}_u+ \mb{X}_l$        & \textbf{81.0 $\pm$ 0.4} & 81.6 $\pm$ 1.1 & 85.9 $\pm$ 0.8   \\
                              & (3) $\mb{X}_u+ (\mb{X}_l, Y_l)$ & 76.2 $\pm$ 0.6          & 80.0 $\pm$ 1.6 & \textbf{87.5 $\pm$ 1.2}            \\
                              \midrule
    \multirow{3}{*}{Best head} & (1) $\mb{X}_u$                 & 71.9 $\pm$ 0.3          & 71.9 $\pm$ 0.3          & 71.9 $\pm$ 0.3            \\
                              & (2) $\mb{X}_u+ \mb{X}_l$        & \textbf{85.3 $\pm$ 0.5} & \textbf{85.2 $\pm$ 1.0} & 89.2 $\pm$ 0.3   \\
                              & (3) $\mb{X}_u+ (\mb{X}_l, Y_l)$ & 77.5 $\pm$ 0.7          & 82.0 $\pm$ 1.6          & 88.3 $\pm$ 1.1            \\
      \bottomrule
    \end{tabular}
\end{table}

\paragraph{Experimental Setup}
We conduct experiments on UNO~\cite{fini2021unified}, which is to the best of our knowledge the current state-of-the-art method in NCD.
To perform experiments (1) and (2), we make adjustments on the framework of UNO, enabling it to run fully self-supervised. 
This is done by replacing the labeled set's ground truth labels $y_{l_{GT}}$ with self-supervised pseudo labels $y_{l_{PL}}$, which are obtained by applying the Sinkhorn-Knopp algorithm~\cite{cuturi2013sinkhorn}.

% Since experiment settings (1) and (2) do not make use of any labels, we need to adapt UNO~\cite{fini2021unified} to work without labeled data.
The standard UNO method conducts NCD in a two-step approach.
In the first step, it applies a supervised pretraining on the labeled data only.
The pretrained model is then used as an initialization for the second step, in which the model is trained jointly on both labeled and unlabeled data using one labeled head and multiple unlabeled heads.
% To achieve this, the logits of known and novel classes are concatenated and the model is trained using a single cross-entropy loss.
% Here, the targets for the unlabeled samples are taken from pseudo-labels, which are generated from the logits of the unlabeled head using the Sinkhorn-Knopp algorithm~\cite{cuturi2013sinkhorn}

To adapt UNO to the fully unsupervised setting in (1), we need to remove all parts that utilize the labeled data. 
Therefore, in the first step, we replace the supervised pretraining by a self-supervised one, which is trained only on the unlabeled data.
For the second step, we simply remove the labeled head, thus the method is degenerated to a clustering approach based solely on the pseudo-labels generated by the Sinkhorn-Knopp algorithm.
For setting (2), we apply the self-supervised pretraining based on both unlabeled and labeled images to obtain the pretrained model in the first step.
In the second step, we replace the ground-truth labels for the known classes with pseudo-labels generated by the Sinkhorn-Knopp algorithm based on the logits of these classes. 
Taken together, the updated setup utilizes the labeled images, but not their labels.

\paragraph{Hyperparameters}
We conduct our experiments on CIFAR100 as well as our proposed ImageNet-based benchmark.
All settings and hyperparameters are kept as close as possible as to the original baselines, including the choice of ResNet18 as the model architecture.
We use SWAV~\cite{caron2020unsupervised} as self-supervised pretraining for all experiments. 
The pretraining is done using the small batch size configuration of the method, which uses a batch size of 256 and a queue size of 3840. 
The training is run for 800 epochs, with the queue being enabled at 60 epochs for our ImageNet-based benchmark and 100 epochs for CIFAR100.
To ensure a fair comparison with the standard NCD setting, the same data augmentations were used.
In the second step of UNO, we train the methods for 500 epochs on CIFAR100 and 100 epochs for each setting on our benchmark.
The experiments are replicated 10 times on CIFAR100 and 5 times on the developed benchmark, and the averaged performances and their corresponding standard errors are summarized in Table \ref{tab:counterintuitive}.

\paragraph{Results}
As suggested in Table \ref{tab:counterintuitive} and Table \ref{tab:counterintuitive u2}, NCD performance is consistently improved by incorporating more images (without labels) from a labeled set, the percentages of improvements in terms of accuracy are around 10\% on CIFAR100 (comparing (1) and (2)). 
% As expected, the additional images are beneficial in all cases, and improve accuracy by around 10\% on CIFAR100 compared to (1).
For our benchmark, the setting (2) obtains an improvement about 6 - 10\% over (1) and
the increase is more obvious in the lower semantic similarity cases.
% with a larger increase for the lower semantic similarity cases.
Similarly, by comparing (2) and (3), we can isolate the impact of the labels. 
For this case, the largest percentage of improvement, about 4 - 7\%, is obtained in the high-similarity setting, followed by the the medium-similarity setting with 1 - 3\% improvements. 
Interestingly, on CIFAR100-50 and ImageNet with low semantic similarity, we unexpectedly observe that (2) performs around 2 - 8\% better than (3), yielding that ``low-quality" supervised information may hurt NCD performance.

\subsection{ {Practical Applications}}
\label{sec: appendix-practical}

As shown in Table \ref{tab:counterintuitive} and Table \ref{tab:counterintuitive u2}, even though we find that supervised knowledge from the labeled set may cause harm rather than gain, it is still difficult to decide whether to utilize supervised knowledge with labeled data or just pure self-supervised knowledge without labels.
Therefore, we offer two concrete solutions, a practical metric (i.e., \pleakage) and a straightforward method.

The proposed \pleakage is a practical reference to infer what sort of data we want to use in NCD,
images-only information, $\mb{X}_u+ \mb{X}_l$ or the image-label pairs, 
$\mb{X}_u+ (\mb{X}_l, Y_l)$ from the labeled set.
% The results from Table \ref{tab:pt-leak} indicate that \pleakage is consistent with the accuracy based on various datasets.
In Table \ref{tab:pt-leak-acc}, we compute \pleakage via a supervised model and a self-supervised model based on pseudo labels. 
As suggested,, the \pleakage is consistent with the accuracy based on various datasets. 
For example, in $L_1-U_1$, the \pleakage computed on the supervised model is 
larger than the one computed in the self-supervised model, 
which is consistent with accuracy, the supervised method outperforms the self-supervised one.  
Reversely, for $L_{2}$-$U_{1}$, $L_{1}$-$U_{2}$ and $L_{1.5}-U_1$, 
the \pleakage computed on the self-supervised model is larger than the one computed in the supervised model, 
which is again consistent with their relative performance.
For $L_2-U_2$ and $L_{1.5}-U_1$, performance in the two settings is within error margins.

% 
% Reversely, for $L_{2}$-$U_{1}$, the pseudo \leakage computed on the self-supervised model is larger than the one computed in the supervised model, which is again consistent with their relative performance.}

\begin{figure}[h]
    \centering
    \begin{tikzpicture}
        \begin{axis}[
            separate axis lines,
            y axis line style={Magenta, line width=1.2pt},
            yticklabels={\textcolor{Magenta}{60}, \textcolor{Magenta}{65}, \textcolor{Magenta}{70}},
            ytick style={Magenta},
            axis x line*=bottom,
            axis y line=left,
            clip=true,
            width={0.5\textwidth},
            xtick={0, 0.25, 0.5, 0.75, 1.0},
            ytick={60, 65, 70},
            y tick label style={font=\small},
            x tick label style={font=\small}, 
            xlabel={$\alpha$\vphantom{Transfer-leakage}},
            ylabel={Accuracy in \%},
            legend columns = 2,
            legend style={at={(0.5,1.0)},anchor=south,fill=none, draw=none},
            xmin=0.0,
            xmax=1.0,
            ymin=59.0,
            ymax=70.5,
            l1u1/.style={mark=triangle*, Magenta},
            l2u1/.style={mark=*, NavyBlue},
            l3u1/.style={mark=square*, Magenta},
            standard uno legend/.style={
              legend image code/.code={
                \draw[##1,/tikz/.cd, dashed, Magenta]
                (0cm,0cm) -- (0.3cm,0cm);
                \draw[##1,/tikz/.cd, dashed, NavyBlue]
                (0.3cm,0cm) -- (0.6cm,0cm);
                },
              },
            ]
            \addplot[forget plot, mark=none, dashed, Magenta, samples=2, opacity=0.75] {62.2};
    
            \addplot [l3u1] table {
                0.00    64.1
                0.25    64.6
                0.50    65.5
                0.70    64.2
                1.00    64.8
                };
            \addlegendentry{\small CIFAR100}
  
            \addlegendimage{mark=*, NavyBlue}
            \addlegendentry{\small ImageNet $L_{2}U_{1}$}
  
            \addlegendimage{standard uno legend}
            \addlegendentry{\makebox[0pt][l]{\hspace{-0.6cm} Standard UNO Accuracy}}
            \addlegendimage{empty legend}
            \addlegendentry{}
        \end{axis}
        \begin{axis}[
          separate axis lines,
          y axis line style={NavyBlue, line width=1.2pt},
          x axis line style={draw=none},
          xmajorticks=false,
          xtick style={draw=none},
          yticklabels={\textcolor{NavyBlue}{75}, \textcolor{NavyBlue}{80}, \textcolor{NavyBlue}{85}},
          ytick style={NavyBlue},
          y tick label style={font=\small}, 
          axis x line*=bottom,
          axis y line=right,
          clip=true,
          width={0.5\textwidth},
          xtick={0, 0.25, 0.5, 0.75, 1.0},
          ytick={75, 80, 85},
          legend columns = 3,
          legend style={at={(0.5,1.0)},anchor=south,fill=none},
          xmin=0.0,
          xmax=1.0,
          ymin=74.5,
          ymax=86,
          l1u1/.style={mark=triangle*, Magenta},
          l2u1/.style={mark=*, NavyBlue},
          l3u1/.style={mark=square*, SpringGreen},
          ]
  
          \addplot [l2u1] table { 
              0.00    80.3
              0.25    82.2
              0.50    81.3
              0.70    80.7
              1.00    79.4
              };
          \addplot[mark=none, dashed, NavyBlue, samples=2, opacity=0.75] {77.2};
      \end{axis}
    \end{tikzpicture}
    \vspace{-\baselineskip}
    \caption{
  %   Experiments on combining supervised and self-supervised objectives, carried out on CIFAR100 and a low-similarity split of our proposed benchmark  (note the different scales).
  %   $\alpha$ indicates the weight of the supervised component.
  %   % The dashed lines show the performance of standard UNO with supervised pretraining.
  %   The results show that for these low-to-medium-similarity cases, a combination of supervised and self-supervised objectives can achieve better results than
  %   either can alone. 
    % Experiments on combining supervised and self-supervised objectives under CIFAR100 and $L_2-U_1$, low semantic case (note the different scales).
    % $\alpha$ indicates the weight of the supervised component.
    %  The dashed lines show the performance of standard UNO with supervised pretraining.
    % The results show that for these low-similarity cases, a combination of supervised and self-supervised objectives can achieve better results than
    % either can alone. 
    Experiments on combining supervised and self-supervised CIFAR100 and $L_2-U_1$, low-similarity setting (note the different scales).
    $\alpha$ shows the weight of the supervised component.
    Dashed lines show the accuracy of SOTA (UNO).
    In low-similarity settings, a mix of supervised and self-supervised objectives outperforms either alone.
    }
    \label{fig:alpha vs acc}
     \vspace{-3mm}
  \end{figure}
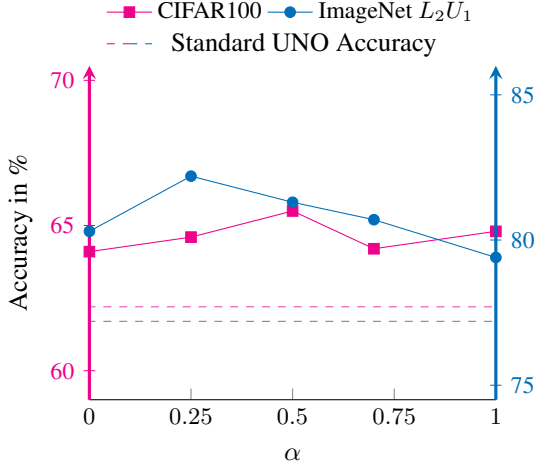

Also, we propose a straightforward method, which smoothly combines supervised knowledge and self-supervised knowledge. 
Concretely, instead of using either the labeled set's ground truth labels $y_{l_{GT}}$ or the self-supervised pseudo labels $y_{l_{PL}}$, we use a linear combination of the two.
The overall classification target is $\alpha y_{l_{GT}} + (1 - \alpha) y_{l_{PL}}$, where $\alpha \in [0, 1]$ is the weight of the supervised component.
This means that for $\alpha = 1$, this approach has the same target as UNO~\cite{fini2021unified}, and differs only in the used pretraining, which is self-supervised as opposed to supervised in UNO.
% Compared to supervised pretraining, self-supervised pretraining achieve 2-3\% improvement in CIFAR100-50 
% and low semantic case ($L_{2}$-$U_{1}$) and competitive performance in other cases as shown in Table \ref{tab:pretrained}.
%we combine the labeled set's ground truth labels $y_{l_{GT}}$ and self-supervised pseudo labels $y_{l_{PL}}$, which are obtained from the self-supervised framework (we exploit SwAV \cite{caron2020unsupervised} here). 
% The overall classification target is $\alpha y_{l_{GT}} + (1 - \alpha) y_{l_{PL}}$, where $\alpha \in [0, 1]$.
As indicated in Figure \ref{fig:alpha vs acc}, our proposed method achieves 3\% and 5\% improvement in both CIFAR100 and ImageNet ($L_{2}$-$U_{1}$), respectively compared to UNO, full results can be found in Table \ref{tab:alpha}.
Our method delivers significant improvements for low semantic similarity cases and competitive performances for high semantic cases.

\begin{table}[H]
    \centering
    \caption{
    {Detailed results on combining supervised and self-supervised objectives. The $\alpha$ value indicates the weight of the supervised component. 
    We see that the low similarity setting sees an improvement of up to 2.8\% compared to $\alpha=1.00$ and up to 5.0\% compared to standard UNO. Standard UNO differs from the $\alpha=1.00$ setting in that it does not use self-supervised pretraining, hence the difference in performance.}
    }
    \label{tab:alpha}
    \begin{tabular}{@{}llccc@{}}
    
          \toprule
    \multirow{2}{*}{Setting}  &\multirow{2}{*}{CIFAR100-50} & \multicolumn{3}{c}{ImageNet $U_{1}$} \\
     \cmidrule(l){3-5} 
     & &  $L_{1}$ - high & $L_{1.5}$ - medium & $L_{2}$ - low\\
    \midrule
    $\alpha$ = 0.00                & 64.1 $\pm$ 0.4          & 79.6 $\pm$ 1.1          & 79.7 $\pm$ 1.0    & 80.3 $\pm$ 0.3          \\
    %$\alpha$ = 0.25             &                                                       & 64.6 $\pm$ 0.6          & 80.9 $\pm$ 1.0          & 80.0 $\pm$ 1.3    & \textbf{82.2 $\pm$ 0.5} \\
    $\alpha$ = 0.50                                                                   & \textbf{65.5 $\pm$ 0.5} & 82.3 $\pm$ 1.6          & 80.2 $\pm$ 1.6    & \textbf{81.3 $\pm$ 1.0}         \\
    %$\alpha$ = 0.75             &                                                       & 64.2 $\pm$ 0.3          & 83.2 $\pm$ 1.5          & 80.7 $\pm$ 1.1    & 80.7 $\pm$ 1.9          \\
    $\alpha$ = 1.00                                                                   & 64.8 $\pm$ 0.8	      & 83.3 $\pm$ 0.6 & 	81.5 $\pm$ 1.0 &	79.4 $\pm$ 0.4 \\
    \midrule
    UNO (SOTA)~\cite{fini2021unified}                                            & 62.2 $\pm$ 0.2          & 83.9 $\pm$ 0.6 & 81.0 $\pm$ 0.6    & 77.2 $\pm$ 0.8\\
    \bottomrule
    \end{tabular}
\end{table}

\begin{table}[H]
        \centering
        \setlength{\tabcolsep}{2.2pt}
        \caption{
        {Comparison of recent NCD methods with our proposed approach which combines supervised and self-supervised objectives.
        }
        }
        \label{tab:comparison-ncd}
        % \vspace{-0.2cm}
        \begin{tabular}{@{}lccccccc@{}}
        
                \toprule
                \multirow{2}{*}{Setting} & \multirow{2}{*}{CIFAR100-50}  & \multicolumn{3}{c}{Unlabeled set $U_{1}$}     & \multicolumn{3}{c}{Unlabeled set $U_{2}$}  \\
                \cmidrule(l){3-5} \cmidrule(l){6-8} 
                & &  $L_{1}$ - high &  $L_{1.5}$ - medium &  $L_{2}$ - low & $L_{1}$ - low & $L_{1.5}$ - medium & $L_{2}$ - high\\
        \midrule
        % $\alpha$ = 0.00  & 64.1 $\pm$ 0.4      & 79.6 $\pm$ 1.1    & 79.7 $\pm$ 1.0    & 80.3 $\pm$ 0.3 & 85.3 $\pm$ 0.5  & 85.2 $\pm$  1.0 & 89.2 $\pm$ 0.3 \\
        % % $\alpha$ = 0.25   & 64.6 $\pm$ 0.6      & 80.9 $\pm$ 1.0          & 80.0 $\pm$ 1.3    & \textbf{82.2 $\pm$ 0.5} & 84.9 $\pm$ 1.4  & 84.9 $\pm$ 1.6  & \textbf{92.2 $\pm$ 0.6} \\
        % $\alpha$ = 0.50            & \textbf{65.5 $\pm$ 0.5} & 82.3 $\pm$ 1.6          & 80.2 $\pm$ 1.6    & \textbf{81.3 $\pm$ 1.0}   &  85.2 $\pm$ 0.9  & \textbf{86.5 $\pm$ 0.6}  & 90.5 $\pm$ 1.2 \\
        % % $\alpha$ = 0.75                  & 64.2 $\pm$ 0.3          & 83.2 $\pm$ 1.5          & 80.7 $\pm$ 1.1    & 80.7 $\pm$ 1.9 & 85.7  $\pm$ 0.1 & 85.5 $\pm$ 1.7  & 92.2 $\pm$ 0.9 \\
        % $\alpha$ = 1.00                  & 64.8 $\pm$ 0.8	      & 83.3 $\pm$ 0.6 & 	81.5 $\pm$ 1.0 &	79.4 $\pm$ 0.4 & 85.8 $\pm$ 0.8  & 85.5 $\pm$ 1.3  & 91.5 $\pm$ 1.1 \\
        % \midrule
        
        RS & 39.2 $\pm$ 2.3	&	55.3 $\pm$ 1.0	&	50.3 $\pm$ 2.0	&	53.6 $\pm$ 1.3	&	48.1 $\pm$ 0.8	&	50.9 $\pm$ 1.3	&	55.8 $\pm$ 1.5 \\
        NCL & 53.4 $\pm$ 0.3	&	75.1 $\pm$ 0.8	&	74.3 $\pm$ 0.4	&	71.6 $\pm$ 0.4	&	61.3 $\pm$ 0.1	&	70.5 $\pm$ 0.8	&	75.1 $\pm$ 1.2	\\
        UNO                                          & 62.2 $\pm$ 0.2          & 83.9 $\pm$ 0.6 & 81.0 $\pm$ 0.6    & 77.2 $\pm$ 0.8 & 77.5 $\pm$ 0.7 & 82.0 $\pm$ 1.6 & 88.3 $\pm$ 1.1 \\
        \midrule
        Ours & \textbf{65.5 $\pm$ 0.5} & 83.3 $\pm$ 0.6 & 81.5 $\pm$ 1.0 & \textbf{81.3 $\pm$ 1.0} & \textbf{85.8 $\pm$ 0.8} & \textbf{86.5 $\pm$ 0.6} & \textbf{91.5 $\pm$ 1.1} \\
        \bottomrule
        \end{tabular}
        \end{table}

\begin{table}[H]
\centering
\caption{
{Comparison of different pretrained models.
Self-supervised pretraining is beneficial for low semantic similarity cases.}
}
\label{tab:pretrained}
\begin{tabular}{@{}lcccc@{}}

        \toprule
\multirow{2}{*}{Pretrained model} 
    & \multirow{2}{*}{CIFAR100-50} & \multicolumn{3}{c}{ImageNet $U_{1}$} \\
    \cmidrule(l){3-5} 
    & & $L_{1}$ - high & $L_{1.5}$ - medium & $L_{2}$ - low\\
\midrule
Self-supervised      & \textbf{64.8 $\pm$ 0.8} & 83.3 $\pm$ 0.6 & 81.5 $\pm$ 1.0 & \textbf{79.4 $\pm$ 0.4} \\ 
Supervised           & 62.2 $\pm$ 0.2                         & 83.9 $\pm$ 0.6                         & 81.0 $\pm$ 0.6                         & 77.2 $\pm$ 0.8                         \\
\bottomrule
\end{tabular}
\end{table}

\clearpage

\section{{Notations}}
To proceed, we summarize all notations used in the paper in Table \ref{tab:notation}.
\begin{table}[h]
    \begin{center}
    \caption{Notation used in the paper.}
    \label{tab:notation}
    \begin{tabular}{p{5cm}l}
    \toprule
      Notation & Description \\
      \midrule
      $\mb{X}_l, \mb{X}_u$ & labeled data / unlabeled data \\
      $y_l, y_u$ & label of labeled data / unlabeled data \\
      $\mathcal{X}_l, \mathcal{X}_u$ & domain of labeled data / unlabeled data \\
      $\mathcal{C}_l, \mathcal{C}_u$ & label set of labeled data / unlabeled data \\
      $\mathbb{P}, \mathbb{Q}$ & probability measure of labeled data / unlabeled data \\
      $\mathcal{L}_n = (\mb{X}_{l,i}, Y_{l,i})_{i=1,\cdots,n}$ & labeled dataset \\
      $\mathcal{U}_m = (\mb{X}_{u,i})_{i=1,\cdots,m}$ & unlabeled dataset \\
      $\mathcal{H}$ & reproducing kernel Hilbert space (RKHS) \\
      $K(\cdot, \cdot)$ & kernel function \\
      $(\mb{X}', Y')$ & independent copy of $(\mb{X}, Y)$ \\
      $ \widehat{\mathbb{P}} $ & estimated probability measure of labeled data \\
      $\mathbb{E}_{\mathbb{Q}}$ & expectation with respect to the probability measure $\mathbb{Q}$ \\
      $\mb{x}_{u,i}, y_{u,i}$ & the $i$-th unlabeled data \\
      $\mathcal{I}_{u,c}$ & index set of unlabeled samples labeled as $y_{u,i} = c$ \\
      $\mb{s}(\mb{x}_{u, i})$ & representation of the $i$-th unlabel data \\
      \bottomrule
    \end{tabular}
    \end{center}
    \end{table}

\section{Related Work }

Novel class discovery (NCD) is a relatively new problem proposed in recent years, aiming to discover novel classes (i.e., assign them to several clusters) by making use of similar but different known classes.
Compared with unsupervised learning, NCD also requires labeled known-class data to help cluster novel-class data. 
NCD is first formalized in DTC \cite{han2019learning}, but the study of NCD can be dated back to earlier works, such as KCL \cite{hsu2018learning} and MCL \cite{hsu2019multi}.
Both of these methods are designed for general task transfer learning, and maintain two models trained with labeled data and unlabeled data respectively.
In contrast, DTC first learns a data embedding on the labeled data with metric learning, then employs a deep embedded clustering method based on \cite{xie2016unsupervised} to cluster the novel-class data.

Later works further deviate from this approach.
Both RS \cite{han2021autonovel,han2020automatically} and \cite{zhao2021novel} use self-supervised learning to boost feature extraction and use the learned features to obtain pairwise similarity estimates.
Additionally, \cite{zhao2021novel} improves on RS by using information from both local and global views, as well as mutual knowledge distillation to promote information exchange and agreement.
NCL \cite{zhong2021neighborhood} extracts and aggregates the pairwise pseudo-labels for the unlabeled data with contrastive learning and generates hard negatives by mixing the labeled and unlabeled data in the feature space.
This idea of mixing labeled and unlabeled data is also used in OpenMix \cite{zhong2021openmix}, which mixes known-class and novel-class data to learn a joint label distribution.
The current state-of-the-art, UNO \cite{fini2021unified}, combines pseudo-labels with ground-truth labels in a unified objective function that enables better  use of synergies between labeled and unlabeled data without requiring self-supervised pretraining.
% Whereas the regular NCD setting is concerned with cases in which both labeled set and unlabeled set are relatively large, \citet{chi2021meta} discuss a setting in which only a few novel-class samples are available. 
Additionally, there are a few theatrical works.
Meta discovery \cite{chi2021meta} indicates that NCD is theoretically solvable if known and unknown classes share high-level semantic features and propose a solution that links NCD to meta-learning.
OSLS \cite{garg2022domain} estimates the target label distribution, including the novel class and learn a target classifier.

\section{Discussion}

The key assumption of novel class discovery is that the knowledge contained in the labeled set can help improve the clustering of the unlabeled set.
Yet, what's the `dark knowledge' transferred from the labeled set to the unlabeled set is still a mystery.
Therefore, we conduct preliminary experiments to disentangle the impact of the different components of the labeled set, i.e., the images-only information and the image-label pairs. 
The results indicate that NCD performance is consistently improved by incorporating more images (without labels) from a labeled set while the supervised knowledge is not always beneficial.
% Moreover, ``low-quality" supervised information may hurt NCD performance. 
Supervised knowledge (obtained from $\mb{X}_l, Y_l$) can provide two types of information: classification rule and robustness.
However, self-supervised information from $\mb{X}_l$ is primarily responsible for enhancing model robustness.
In cases of high semantic similarity, the labeled and unlabeled classification rules are more similar than in cases of low semantic similarity.
Thus, supervised knowledge is advantageous in high similarity cases but potentially harmful in low similarity situations while self-supervised knowledge is helpful for both cases.

\clearpage

\section{Detailed Benchmark Splits}
\label{sec:appendix-benchmark-splits}
\begin{table}[!h]
    \caption{ImageNet class list of labeled split $L_{1}$ and unlabeled split $U_{1}$ of our proposed benchmark. As they share the same superclasses, they are highly related semantically. For each superclass, six classes are assigned to the labeled set and two to the unlabeled set. The labeled classes marked by the red box are also included in $L_{1.5}$, which shares half of its classes with $L_{1}$ and half with $L_{2}$.}
    \begin{tikzpicture}
        \node (table) {\begin{tabular}{@{}p{\dimexpr 0.2\linewidth-2\tabcolsep}p{\dimexpr 0.5\linewidth-2\tabcolsep}p{\dimexpr 0.3\linewidth-2\tabcolsep}@{}}
    \toprule
    Superclass & Labeled Subclasses & Unlabeled Subclasses \\
    \midrule
    garment & vestment, jean, academic gown, sarong, fur coat, apron & swimming trunks, miniskirt \\
    \midrule
    tableware & wine bottle, goblet, mixing bowl, coffee mug, water bottle, water jug & plate, beer glass\\
    \midrule
    insect & leafhopper, long-horned beetle, lacewing, dung beetle, sulphur butterfly, fly & admiral, grasshopper\\
    \midrule
    vessel & wreck, liner, container ship, catamaran, trimaran, lifeboat & yawl, aircraft carrier\\
    \midrule
    building & toyshop, grocery store, bookshop, palace, butcher shop, castle & beacon, mosque\\
    \midrule
    headdress & cowboy hat, bathing cap, pickelhaube, bearskin, bonnet, hair slide & crash helmet, shower cap\\
    \midrule
    kitchen utensil & cocktail shaker, frying pan, measuring cup, tray, spatula, cleaver & caldron, coffeepot\\
    \midrule
    footwear & knee pad, sandal, clog, cowboy boot, running shoe, Loafer & Christmas stocking, maillot\\
    \midrule
    neckwear & stole, necklace, feather boa, bow tie, Windsor tie, neck brace & bolo tie, bib\\
    \midrule
    bony fish & puffer, sturgeon, coho, eel, rock beauty, tench & gar, lionfish\\
    \midrule
    tool & screwdriver, fountain pen, quill, shovel, screw, combination lock & torch, padlock\\
    \midrule
    vegetable & spaghetti squash, cauliflower, zucchini, acorn squash, artichoke, cucumber & cardoon, butternut squash\\
    \midrule
    motor vehicle & beach wagon, trailer truck, limousine, police van, convertible, school bus & garbage truck, moped\\
    \midrule
    sports equipment & balance beam, rugby ball, ski, horizontal bar, racket, dumbbell & tennis ball, croquet ball\\
    \midrule
    carnivore & otterhound, flat-coated retriever, Italian greyhound, Shih-Tzu, basenji, black-footed ferret & Boston bull, Bedlington terrier \\
    \bottomrule
\end{tabular}
};
        \node [coordinate] (north east) at ($(table.north west) !.7! (table.north east)$) {};
        \node [coordinate] (south east) at ($(table.south west) !.7! (table.south east)$) {};
        \node [coordinate] (bottom right) at ($(north east) !.5325! (south east)$) {};
        \node [coordinate] (top right) at ($(north east) !.05! (south east)$) {};
        \node [coordinate] (bottom left) at ($(north east) !.5325! (south east)$) {};
        \node [coordinate] (top left) at ($(table.north west) !.05! (table.south west)$) {};
        \node [coordinate] (bottom left) at ($(table.north west) !.5325! (table.south west)$) {};
        \node [coordinate] (middle bottom) at ($(bottom left) !.65! (bottom right)$) {};
        \node [coordinate] (middle north) at ($(table.north west) !.65! (top right)$) {};
        \node [coordinate] (middle top) at ($(middle bottom) !.1! (middle north)$) {};
        
        \draw [red,ultra thick,rounded corners]
            (bottom left) -- (top left) -- (top right) |- (middle top) -- (middle bottom) -- (bottom left);
    \end{tikzpicture}
\end{table}

\begin{table}[!h]
    \caption{ImageNet class list of labeled split $L_{2}$ and unlabeled split $U_{2}$ of our proposed benchmark. As they share the same superclasses, they are highly related semantically. For each superclass, six classes are assigned to the labeled set and two to the unlabeled set. The labeled classes marked by the red box are also included in $L_{1.5}$, which shares half of its classes with $L_{1}$ and half with $L_{2}$.}
    \begin{tikzpicture}
        \node (table) {\begin{tabular}{@{}p{\dimexpr 0.2\linewidth-2\tabcolsep}p{\dimexpr 0.5\linewidth-2\tabcolsep}p{\dimexpr 0.3\linewidth-2\tabcolsep}@{}}
    \toprule
    Superclass & Labeled Subclasses & Unlabeled Subclasses \\
    \midrule
    fruit & corn, buckeye, strawberry, pear, Granny Smith, pineapple & acorn, jackfruit\\
    \midrule
    saurian & African chameleon, Komodo dragon, alligator lizard, agama, green lizard, Gila monster & banded gecko, American chameleon\\
    \midrule
    barrier & stone wall, chainlink fence, breakwater, dam, bannister, picket fence & worm fence, turnstile\\
    \midrule
    electronic equipment & cassette player, modem, printer, monitor, computer keyboard, pay-phone & dial telephone, microphone\\
    \midrule
    serpentes & green snake, boa constrictor, green mamba, ringneck snake, thunder snake, king snake & rock python, garter snake\\
    \midrule
    dish & hot pot, burrito, potpie, meat loaf, cheeseburger, mashed potato & hotdog, pizza\\
    \midrule
    home appliance & espresso maker, toaster, washer, space heater, vacuum, microwave & dishwasher, Crock Pot\\
    \midrule
    measuring instrument & wall clock, barometer, digital watch, hourglass, magnetic compass, analog clock & digital clock, parking meter\\
    \midrule
    primate & indri, siamang, baboon, capuchin, chimpanzee, howler monkey & patas, Madagascar cat\\
    \midrule
    crustacean & rock crab, king crab, crayfish, American lobster, Dungeness crab, spiny lobster & fiddler crab, hermit crab\\
    \midrule
    musical instrument & organ, acoustic guitar, French horn, electric guitar, upright, maraca & violin, grand piano\\
    \midrule
    arachnid & black and gold garden spider, wolf spider, harvestman, tick, black widow, barn spider & tarantula, scorpion\\
    \midrule
    aquatic bird & dowitcher, goose, albatross, limpkin, white stork, red-backed sandpiper & drake, crane\\
    \midrule
    ungulate & hippopotamus, hog, llama, hartebeest, ox, gazelle & warthog, zebra\\
    \midrule
    passerine & house finch, magpie, goldfinch, indigo bunting, chickadee, brambling & bulbul, water ouzel\\
    \bottomrule
\end{tabular}
};
        \node [coordinate] (north east) at ($(table.north west) !.7! (table.north east)$) {};
        \node [coordinate] (south east) at ($(table.south west) !.7! (table.south east)$) {};
        \node [coordinate] (bottom right) at ($(north east) !.52! (south east)$) {};
        \node [coordinate] (top right) at ($(north east) !.05! (south east)$) {};
        \node [coordinate] (bottom left) at ($(north east) !.52! (south east)$) {};
        \node [coordinate] (top left) at ($(table.north west) !.05! (table.south west)$) {};
        \node [coordinate] (bottom left) at ($(table.north west) !.52! (table.south west)$) {};
        \node [coordinate] (middle bottom) at ($(bottom left) !.83! (bottom right)$) {};
        \node [coordinate] (middle north) at ($(table.north west) !.83! (top right)$) {};
        \node [coordinate] (middle top) at ($(middle bottom) !.1! (middle north)$) {};

        \draw [red,ultra thick,rounded corners]
            (bottom left) -- (top left) -- (top right) |- (middle top) -- (middle bottom) -- (bottom left);
    \end{tikzpicture}
\end{table}

\begin{table}[!t]
    \centering
    \caption{Labeled Split of CIFAR100 used in Section \ref{sec:benchmark}.
    We construct data settings based on its hierarchical class structure.
    $U_{1}$-$L_{1}$/$U_{2}$-$L_{2}$ are share the same superclasses.}
    \begin{tabular}{@{}p{\dimexpr 0.35\linewidth-2\tabcolsep}p{\dimexpr 0.5\linewidth-2\tabcolsep}p{\dimexpr 0.18\linewidth-2\tabcolsep}@{}}
        \toprule
        Superclass & Labeled Subclasses ($L_{1}$)  & Unlabeled Subclasses ($U_{1}$) \\
        \midrule
        aquatic\_mammals & dolphin, otter, seal, whale & beaver \\
        \midrule
% 		 \vspace{-0.3cm}
% 		&\\
% 		\cdashline{1-3}
% 		\vspace{-0.35cm}
% 		&\\
       fish &  flatfish,  ray,  shark,  trout  &  aquarium\_fish \\
      \midrule
       flower & poppy,  rose,  sunflower,  tulip &  orchids\\
      \midrule
       food containers & bowl,  can,  cup,  plate & bottles\\
       \midrule
       fruit and vegetables & mushroom,  orange,  pear,  sweet\_pepper & apples\\
       \midrule
       household electrical devices & keyboard,  lamp,  telephone,  television & clock\\
       \midrule
       household furniture &  chair,  couch,  table,  wardrobe  & bed\\
       \midrule
       insects & beetle,  butterfly,  caterpillar,  cockroach  & bee\\
       \midrule
       large carnivores &  leopard,  lion,  tiger,  wolf & bear\\
       \midrule
       large man-made outdoor things &  castle,  house,  road,  skyscraper & bridge\\
       \midrule
       \midrule
        Superclass & Labeled Subclasses ($L_{2}$)  & Unlabeled Subclasses ($U_{2}$) \\
        \midrule
        large natural outdoor scenes & forest,  mountain,  plain,  sea & cloud\\
        \midrule
        large omnivores and herbivores & cattle,  chimpanzee,  elephant,  kangaroo & camel\\
        \midrule
        medium-sized mammals& porcupine,  possum,  raccoon,  skunk & fox\\
        \midrule
        non-insect invertebrates & lobster,  snail,  spider,  worm & crab\\
        \midrule
        people & boy,  girl,  man,  woman & baby\\
        \midrule
        reptiles & dinosaur,  lizard,  snake ,  turtle & crocodile\\
        \midrule
        small mammals & mouse,  rabbit,  shrew,  squirrel & hamster\\
        \midrule
        trees & oak\_tree,  palm\_tree,  pine\_tree,  willow\_tree & maple\\
        \midrule
        vehicles 1 & bus,  motorcycle,  pickup\_truck,  train & bicycle\\
        \midrule
        vehicles 2 & rocket,  streetcar,  tank,  tractor & lawn-mower\\
        
         \bottomrule
    \end{tabular}
\end{table}

\clearpage

\end{document}